\documentclass{amsart}
\usepackage{subfigure}
\usepackage{xcolor}
\usepackage{amsthm}
\usepackage{amsaddr}
\usepackage{csquotes}
\usepackage{caption}
\usepackage{mathtools}
\usepackage{graphicx,tikz}
\usetikzlibrary{shapes,arrows}
\usetikzlibrary{calc, positioning, shapes, through, intersections}
\usepackage{graphicx}
\usepackage{wrapfig}
\usepackage{comment}
\newcommand{\rev}{}

\newcommand{\tr}{\rm Tr}
\newcommand\numberthis{\addtocounter{equation}{1}\tag{\theequation}}

\newcommand{\utheta}{\omega}
\newcommand{\uomega}{\omega}
\newcommand{\ux}{x}

\usepackage{xr-hyper}
\makeatletter
\newcommand*{\addFileDependency}[1]{
  \typeout{(#1)}
  \@addtofilelist{#1}
  \IfFileExists{#1}{}{\typeout{No file #1.}}
}
\makeatother

\newcommand*{\myexternaldocument}[1]{%
    \externaldocument{#1}%
    \addFileDependency{#1.tex}%
    \addFileDependency{#1.aux}%
}
\myexternaldocument{response}

\newcommand{\xMapsto}[2][]{\ext@arrow 0599{\Mapstofill@}{#1}{#2}}
\def\Mapstofill@{\arrowfill@{\Mapstochar\Relbar}\Relbar\Rightarrow}

\newtheorem*{problem*}{Problem}
\usepackage{algorithmic}

\usepackage{graphicx,colordvi}
\usepackage{amsmath,amssymb}
\usepackage{epstopdf}
\usepackage[caption=false]{subfig}
\usepackage{epsfig,cite}
\usepackage{nicefrac}
\usepackage{enumerate}
\usepackage{algorithmic}
\usepackage[algoruled,vlined]{algorithm2e}

\newtheorem{remark}{Remark}

\newtheorem{lemma}{Lemma}

\newcommand{\MYfooter}{\smash{
		\hfil\parbox[t][\height][t]{\textwidth}{
		}\hfil\hbox{}}}
\makeatletter
\def\ps@IEEEtitlepagestyle{%
	\def\@oddfoot{\MYfooter}%
	\def\@evenfoot{\MYfooter}}
\makeatother
\def\trace{\mathop{\rm Tr}\nolimits}

\begin{document}

\title[parameter estimation with unknown observation]{Kernel-based parameter estimation of dynamical systems with unknown observation functions}

\author[]{Ofir Lindenbaum*}
\address{Program in Applied Mathematics, Yale University, 51 Prospect Street, New Haven, CT 06511,~USA}
\thanks{*OL and AS contributed equally to this work.}
\email{ofir.lindenbaum@yale.edu}

\author[]
{Amir Sagiv*}
\address{Department of Applied Physics and Applied Mathematics, Columbia University, 500~West 120th Street, New York, NY 10027, USA}
\email{as6011@columbia.edu}

\author[]{Gal Mishne}
\address{Halicioglu Data Science Institute, UC San Diego
9500 Gilman Drive MS 0555
SDSC 215E
La Jolla, CA  92093-0555}
\email{gmishne@ucsd.edu}

\author[]{Ronen Talmon}
\address{Faculty of Electrical Engineering,
Technion - Israel Institute of Technology,
Haifa 32000, Israel}
\email{ronen@ef.technion.ac.il}

\maketitle
\begin{abstract}
    A low-dimensional dynamical system is observed in an experiment as a high-dimensional signal; for example, a video of a chaotic pendulums system. Assuming that we know the dynamical model up to some unknown parameters, can we estimate the underlying system's parameters by measuring its time-evolution only once? The key information for performing this estimation lies in the temporal inter-dependencies between the signal and the model. We propose a kernel-based score to compare these dependencies. Our score generalizes a maximum likelihood estimator for a linear model to a general nonlinear setting in an unknown feature space. We estimate the system's underlying parameters by maximizing the proposed score. We demonstrate the accuracy and efficiency of the method using two chaotic dynamical systems - the double pendulum and the Lorenz '63 model.
\end{abstract}

\section*{Lead Paragraph}

The purpose of many experimental designs is to measure a quantity of interest by observing a dynamical system and comparing the observations to a known model. This procedure can be difficult when the system is chaotic and can be even more challenging when the correspondence between the measurements and the model, the observation function, is unknown. For a single experiment, i.e., when the observed data is a single time-series, learning the unknown observation function is not an option. Instead, we construct a kernel-based score in a way that is agnostic to the unknown observation function. We can derive a maximum likelihood estimator for identifying the observed dynamical system's parameters by maximizing that score. The intuition behind our approach is that even though the map between the coordinates/model and the observations is unknown, the dynamics of the data convey enough information on the dynamics of the model with the true parameter, thus facilitating an informed parameter estimation procedure. We propose two optimization schemes for maximizing our score. Finally, we demonstrate that our method can accurately estimate the governing parameters for two chaotic dynamical systems from complex and high-dimensional data.




\section{Introduction}

Consider a common situation in experimental sciences - an experiment is designed to measure a quantity of interest by observing a dynamical system and comparing the observations to a known model. But can this measurement be performed when the model is complex and chaotic? Furthermore, is this procedure possible when the correspondence between the measurement and the model, the {\em observation function}, is unknown?

    \begin{figure}[h!]
    \includegraphics[width=1\textwidth]{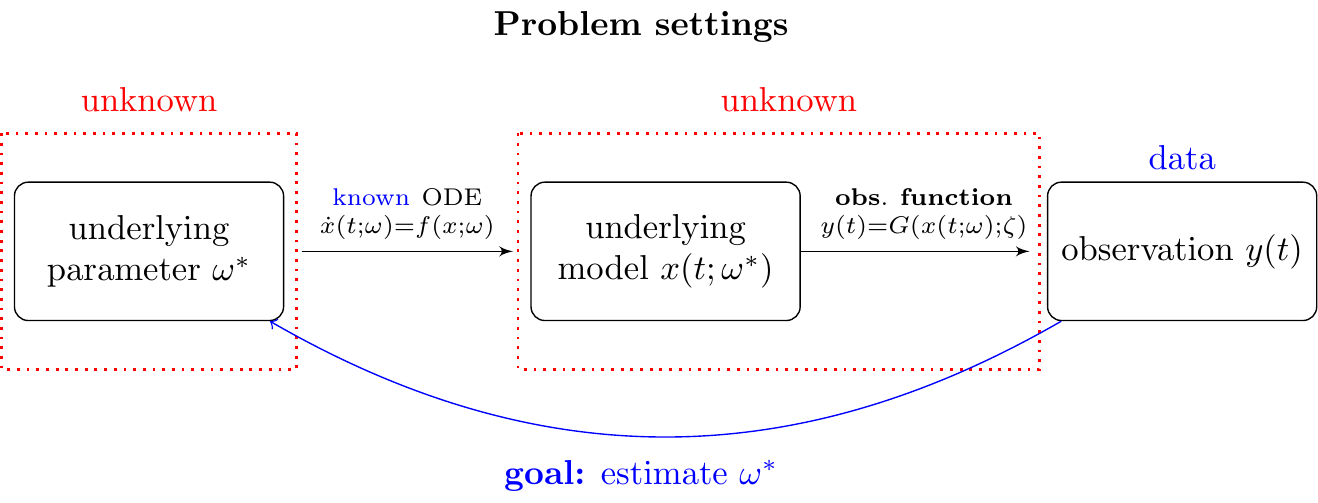}

\caption{The schematic settings of our problem. We are given the observations $y(t)$ and a mechanism to generate $x(t;\omega)$ for every $\omega\in \Omega$ (an ODE). What is the true underlying parameter $\omega^*$ driving~$y(t)$?}
\label{fig:problem}
\end{figure}

  
For example, it is straightforward to estimate the gravitational free acceleration $g$ by observing a pendulum; the angle $x(t)$ of a pendulum of length $\ell$ varies periodically according to the harmonic oscillator ordinary differential equation (ODE) $\ddot{x}(t)=-(g/\ell)x$. By solving the ODE, $g$ may be estimated using $g=\nu ^2 \ell$, where the frequency $\nu$ can be directly observed from the pendulum's oscillations. But can such a measurement scheme be applied to the chaotic {\em double} pendulum, where no easily observable parameter like the frequency $\omega$ exists?

The double pendulum example illustrates a more general class of problems; see Fig.\ \ref{fig:problem}. In an experiment, an observed signal $y(t)$ is related to its governing model $x(t;\omega ^*)$ by specific yet unknown parameters (or parameter-vector) $\omega^*$ and an unknown {\em observation function} $G$, i.e., $$y(t)=G(x(t;\omega^*);\zeta) \, ,$$ where $\zeta$ is a noise source. The purpose of this study is to estimate the system's parameters $\omega^*$ among all possible parameters $\omega$ in a parameter space $\Omega$, using the observation $y(t)$ and the general model $x(t;\omega)$. Critically, we note that even though the map $\omega \mapsto x(t;\omega)$ is known, the map $ x\xmapsto[]{G} y$ is {\em unknown} to us. Therefore, we only know ``half" of the forward map $\omega\mapsto y(t)$. Since the forward map is unknown, this problem does not fit into the usual definition of inverse problems \cite{aster2018parameter, stuart2010inverse}. Conversely, since we only observe a single experiment and do not have a lot of data, it is not straightforwardly amenable to standard machine learning methodology (see Sec.\ \ref{sec:machlearn} for details). 

    \begin{figure}[h!]
    \includegraphics[width=1\textwidth]{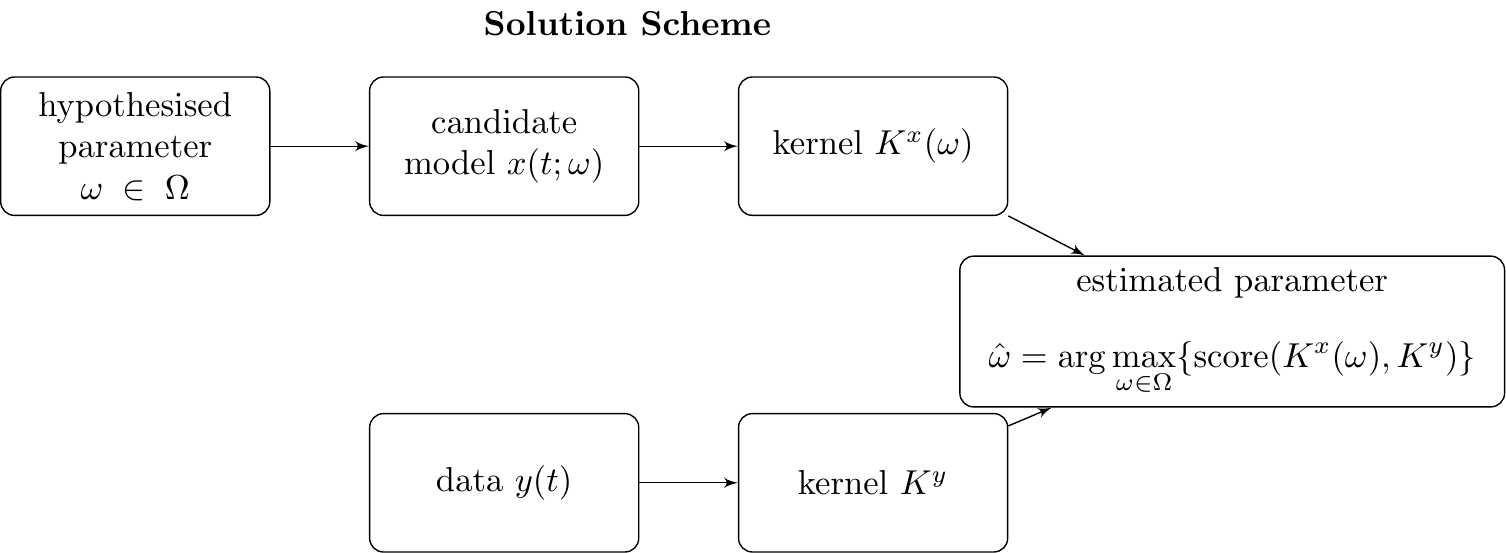}
\caption{The schematics of the proposed solution. For every~$\omega\in \Omega$ a kernel $K^x(\omega)$ is computed. This Kernel is compared to the observation kernel $K^y$, and the estimated $\hat{\omega}$ is chosen to maximize their similarity score. The hypothesised $\omega$ values are either predetermined (Algorithm \ref{alg:search}) or dynamically determined using an optimization scheme (Algorithm \ref{alg:opt}).}
\label{fig:solution}
\end{figure}
We propose a kernel-based approach to estimate the system's parameters~$\omega^*$. We first study the case of a linear observation function $G$. A maximum-likelihood estimation of $\omega^*$ then yields a maximization problem for a normalized variant of the cross-covariance between the observations and the model. To carry this idea to the general nonlinear case, we ``lift" both the observations and the model to an infinite-dimensional Hilbert space (feature space \cite{hofmann2008kernel, sahwe2004kernel}). In the feature space, the two signals are again linearly dependent. By constructing kernels for the observations $y(t)$ and for the model $x(t;\omega)$, a covariance-like score in the feature space is computed (see \eqref{eq:score}) and maximized to estimate the system's parameters. By applying our method (Algorithms \ref{alg:search} and \ref{alg:opt}) to two examples of chaotic dynamical systems - the double pendulum and the Lorenz system - we demonstrate empirically that maximizing the kernel-based score indeed yields an accurate estimate for $\omega^*$.

The application of the so-called kernel trick to generalize the linear notion of covariance has been used for various statistical tasks such as kernel principle component analysis (PCA), kernel canonical-correlation analysis (CCA), and the Hilbert-Schmidt Independence criteria \cite{bach2002ica, fukumizu2007cca, gretton2005measure, michaeli2016cca, scholkopf1998kernel}. Kernels were also used in this context of kernel density estimators \cite{vestner2017product} or for extracting latent variables from multi-modal observations as in \cite{lindenbaum2017multi, lederman2018diff,salhov,yair}. While resembling to some nonlinear kernel statistical problems on the one hand, and to some machine learning and model discovery problems on the other hand \cite{atkinson2019data, bertalan2020learning, bramburger2020poincre, dsilva2018chemo, ho2020discover, kutz2016pnas, kutz2019data, yair2017reconstruction}, we note that the problem of parameter learning under an unknown observation function is stated here, to the best of our knowledge, for the first time. Consequently, our kernel-based score does not seem to appear in the kernel methods literature; see discussion in Section \ref{sec:analysis}.
 
 
 The remainder of this paper is organized as follows. Sec.\ \ref{sec:problem} presents the problem in formal terms. In Sec.\ \ref{sec:analysis} we derive our method initially for the linear case, and then to the general nonlinear case. Sec.\ \ref{sec:method} presents the main algorithms of this paper - the search-based Algorithm \ref{alg:search} and the optimization-based Algorithm \ref{alg:opt}. The applications of our approach to the double pendulum and to the Lorenz system are presented in Sec.\ \ref{sec:results}. Finally, we discuss potential applications of the method and its relationship to previous studies on model discovery, inverse problems, and kernel methods in Sec.~\ref{sec:discussion}.

\section{Problem Formulation}\label{sec:problem}
Consider a parametric family of autonomous ordinary differential equations~(ODE)
\begin{equation}
\label{eq:ode} 
\left\{ \begin{array}{l} 
\dot{x}(t;\uomega ) = f(x;\uomega) \, , \qquad \omega\in\Omega \subseteq \mathbb{R}^m \, , \\ x(0;\omega) = x_0 (\uomega)\in \mathbb{R}^d \, , \end{array}\right.
\end{equation}
where $\Omega\subseteq \mathbb{R}^m$ is a convex set of possible parameters and $f$ is sufficiently smooth such solutions are unique and exist globally. The dynamics $x(t;\omega)$ are therefore completely determined by a fixed vector of parameters $\omega ^* \in \Omega$. Assume that $\omega ^*$ is unknown and that we do not observe $x(t;\omega^*)$, but only a measurement $y(t) \in \mathbb{R}^D$ for some dimension $D$. This observation can be viewed as a noisy lifting of $x(t;\omega^*)$ from the latent space $\mathbb{R}^d$ to the ambient observation space $\mathbb{R}^D$ by an unknown and possibly noisy map, i.e., {\rev  \begin{equation}\label{eq:Gdef}
    y(t) =G(x(t;\uomega^*);  \zeta)\, , \qquad G: \mathbb{R}^d\times  \mathcal{Z} \to \mathbb{R}^D \, ,
\end{equation}
where $\mathcal{Z}$ is some manifold} in which $\zeta(t)$ is a stationary random process with $\delta$ auto-correlation and {\rev the observation function $G(\cdot, 0)$ guarantees identifiability of $\omega$; see \cite{villaverde2019obs} and Section \ref{sec:degen} for details} .\footnote{It is not essential that $G$ is defined on all of $\mathbb{R}^d$, but just on $\cup_{\omega \in \Omega} \, \{~\ux (t;\uomega)~ | ~ t\geq 0 \}$. {\rev Furthermore, in our analysis and examples usually $\mathcal{Z}$ is either the space of the model $\mathbb{R}^d$ or the observation space $\mathbb{R}^D$.}}  For example, if $x(t;\omega)$ describes the trajectory of a ballistic projectile in $\mathbb{R}^3$, its video will embed this trajectory in $\mathbb{R}^D$, where $D$ is the number of pixels in each video frame. As a practical matter, we will further assume that $y(t)$ is measured in discrete times $\{t_j = (j-1)\Delta t\}_{j=1}^N$ for some~$\Delta t > 0$. The main problem of this paper can now be formally stated:
\begin{problem*}
 Given 
\begin{enumerate}
    \item A single observed time series $\{y(t_j)\}_{j=1}^N$, defined by \eqref{eq:Gdef} with unknown $G$ and $\omega^*$, and
    \item A solution $x(t;\omega)$ to the ODE \eqref{eq:ode} for all $t\geq 0 $ and $\omega \in \Omega$,
\end{enumerate}
find the vector of underlying parameters $\omega^*$.
\end{problem*}

\begin{remark}
Uncertainty in the observations may stem from several sources -- modelling misspecification, numerical errors, measurement noise, nuisance variables in the experiment etc. The introduction of randomness in \eqref{eq:Gdef} is a modelling decision aimed to capture all of these uncertainty sources.
\end{remark}
We start by considering the simplified linear variant of \eqref{eq:Gdef}, in which we can derive a maximum likelihood estimator of $\omega^*$. In the general nonlinear case, we use the kernel trick to map the phase-space coordinates $x(t;\omega)$ and the observations~$y(t)$~into a Hilbert space (feature space) where the linear approach can be employed again.
\section{Derivation}\label{sec:analysis}
\subsection{The linear case - a maximum likelihood approach}\label{sec:linear}
It is instructive to first consider \eqref{eq:Gdef} where $G$ is linear in $\ux$ and additive with respect to a Gaussian noise term, i.e.,
\begin{equation}\label{eq:linear}
\bar{y}(t_j)=A \bar{x}(t_j;\omega^* )+\zeta_j \, , \qquad  j=1,...,N \, ,  
\end{equation}
where $\bar{x}(t;\omega ) = x(t;\omega )-\mathbb{E}_{\tau}x(\tau ;\omega)$ and $\bar{y}(t) = y(t)-\mathbb{E}_{\tau}y(\tau)$ are centered, $A\in M_{D,d}(\mathbb{R})$, and for each $1\leq j \leq N$ the term $\zeta_j$ is drawn iid from $\mathcal{N}(0,\sigma^2 {\rm I} )$ for some $\sigma >0$.

\begin{remark}
We center the observations $y$ and model coordinates $x$ since the choice of origin in either $\mathbb{R}^D$ and $\mathbb{R}^d$ is arbitrary from a modelling/physics perspective. See more on the role of such invariances in Section \ref{sec:degen}. In practice, the time-averages should be replaced by their empirical estimates, e.g., $\mathbb{E}_{\tau}y(\tau )\approx N^{-1}\sum_j y(t_j)$.
\end{remark}

To estimate $\uomega ^*$ from the observations $\{y(t_n)\}_{n=1}^N$, we use a maximum likelihood (ML) estimator \cite{abramovich2013stat}. The ML estimator is defined as
$$\hat{\omega}_{\rm ml} = \arg \max \limits_{\substack{\uomega \in \Omega\\ {\bf A}\in M_{D,d}(\mathbb{R})}} {\rm Prob}_{\zeta_1, \ldots, \zeta_N}(\bar{y}_1, \ldots, \bar{y}_N | \omega,A)  \, .$$ 
Since the $\rm log$ function is monotonic increasing, we can replace the likelihood function by log-likelihood to exploit the independence of the normal $\zeta_j$'s to get
\begin{align*}
\hat{\uomega} _{\rm ml} &= \arg \max \limits_{\substack{\uomega \in \Omega\\ {\bf A}\in M_{D,d}(\mathbb{R})}} \log {\rm Prob}_{\zeta_1, \ldots, \zeta_N}(\bar{y}_1, \ldots, \bar{y}_N | \omega,A)\\
&=\arg \max\limits_{\substack{\uomega \in \Omega\\ {\bf A}\in M_{D,d}(\mathbb{R})}}-\frac{\sum\limits^N_{n=1}\|\bar{y}(t_n)-A \bar{x}(t_n,\omega)\|_2^2}{2\sigma^2}-\frac{DN}{2}\log(2\pi \sigma^2) \, .  \numberthis \label{eq:ls}
\end{align*}
Since $D$, $N$, $\sigma$, and the observations $y(t_j)$ are independent of $\omega$ and $A$, we can simplify the objective function on the right-hand-side as follows:
\begin{align*}
\hat{\uomega}_{\rm ml} &= \arg \min _{\substack{\uomega \in \Omega\\ A\in M_{D,d}(\mathbb{R})}} \sum^N_{n=1}\|\bar{y}(t_n)-A \bar{x}(t_n,\omega)\|_2^2 \\  
&= \arg \min _{\substack{\uomega \in \Omega\\ A\in M_{D,d}(\mathbb{R})}} \|AX(\uomega)\|_F^2 -2\langle AX(\uomega),Y\rangle + \|Y\|_F^2  \\
&= \arg \min _{\substack{\uomega \in \Omega\\ A\in M_{D,d}(\mathbb{R})}} \|AX(\uomega)\|_F^2 -2\langle AX(\uomega),Y\rangle \, , \numberthis \label{eq:ml_preO}
\end{align*}
where $X(\omega)$ and $Y$ are matrices whose $j$-th columns are $\bar{x}(t_j;\uomega) $ and $\bar{y}(t_j)$, respectively, $\langle B,C\rangle = \sum_{i,j} B_{i,j}C_{i,j}$ is the Frobenius inner product on $M_{D,d}(\mathbb{R})$ and $\|B\|_F =\langle B, B\rangle ^{1/2}$ is the Frobenius norm. Next, we show that one can also normalize \eqref{eq:ml_preO} by $\|AX\|_F$. To see that, fix $\uomega\in \Omega$ and $O\in M_{D,d}(\mathbb{R})$ such that $A= \lambda O$, $\|OX\|_F =1$, and $\lambda \in \mathbb{R}$. Then, by direct differentiation in $\lambda$ $$\|AX(\uomega)\|_F^2 -2\langle AX(\uomega),Y\rangle = \lambda ^2 -2\lambda \langle OX(\uomega),Y\rangle \, , $$
 is minimized when $\lambda = \langle OX, Y\rangle$. Hence, 
$$\hat{\uomega}_{\rm ml} =  \arg \min_{\substack{\uomega \in \Omega\\ \| OX\|_F = 1}}  \langle OX(\uomega) , Y\rangle ^2 - 2\langle \langle OX(\uomega) , Y\rangle OX(\uomega), Y \rangle  \, . $$
By setting $O= A\|AX\|_F^{-1}$, we get that the maximum likelihood estimator is
\begin{equation}\label{eq:ml_ip}
    \hat{\uomega}_{ml}  = \arg \max_{\substack{\uomega \in \Omega \\ A\in M_{D,d}(\mathbb{R})}} ~ \frac{{\Large{<}}AX(\omega), Y {\Large{>}_F}^2 }{\|AX(\omega)\|_F^2\cdot \|Y\|_F^2 } \, ,
\end{equation}
where, since $Y$ is a constant matrix, we divide by $\|Y\|_F^2$ so that the argument on the right hand side is always $\leq 1$.

\subsection{Maximum likelihood in the nonlinear settings - a kernel approach}\label{sec:nl}
In the general model \eqref{eq:Gdef}, the observations $y(t)$ do not depend linearly on the model coordinates $x(t)$ as in \eqref{eq:linear}. Rather, the two depend nonlinearly via $G$, see \eqref{eq:Gdef}. If we knew $G$, the linear maximum-likelihood approach \eqref{eq:ml_ip} could be applied to $y$ and $G(x(\cdot ;\omega^*);\zeta)$ in $\mathbb{R}^D$. Even though we do not know $G$, the relation $\eqref{eq:Gdef}$ implies that~$x$ and $y$ are linearly dependent under {\rev nonlinear transformations $\psi:\mathbb{R}^d\to \mathbb{R}^D$ and $\phi:\mathbb{R}^D \to \mathbb{R}^D$ (feature maps),} respectively, i.e.,
\begin{equation}\label{eq:feature_model}
\phi(y(t)) = \psi(x(t;\omega^*);\zeta)  \, .
\end{equation}
In what follows, we assume that the noise term $\zeta$ is again Gaussian and additive, i.e., $\phi(y(t)) = \psi(x(t;\omega^*))+\zeta $ where $\zeta \sim \mathcal{N}(0,\sigma^2{\rm I})$. {\rev In this case, one possible set of feature maps is simply $\phi(y) = y$ and $\psi(x) = G(x)$.} However, we show in the numerical experiments that our algorithm can estimate $\omega^*$ even for non-Gaussian and non-additive noise sources, see Sec.\ \ref{sec:lorenz}.

\begin{remark}\label{rem:feature}
    Seemingly, the nonlinear model \eqref{eq:feature_model} is more restrictive than its linear counterpart \eqref{eq:linear}, absent of the freedom to choose the linear transformation $A$. Given the maps $\phi$ and $\psi$, however, $A$ is ``absorbed" into the definitions of $\phi$ and $\psi$.
\end{remark}
Using the same maximum-likelihood argument of Sec.\ \ref{sec:linear}, the nonlinear model~\eqref{eq:feature_model} yields the following estimator of $\omega^*$ (compare with \eqref{eq:ml_ip})
\begin{subequations}\label{eq:ml_nl_raw}
\begin{equation}\label{eq:ml_nl_raw_a}
      \hat{\uomega}  = \arg \max_{\substack{\uomega \in \Omega}} ~ \frac{\langle\bar{\Psi}(\omega), \bar{\Phi} \rangle^2_F }{\|\bar{\Psi}(\omega)\|_F^2\cdot \|\bar{\Phi}\|_F^2 } \, ,
\end{equation}
where
\begin{equation}
   \bar{\Psi}_{\cdot, j}(\omega) :\,= \psi(x(t_j;\omega))- \frac{1}{N}\sum\limits_{i=1}^N \psi(x(t_i;\omega)) \, , \qquad  \bar{\Phi}_{\cdot, j} :\,= \phi(y(t_j))- \frac{1}{N} \sum\limits_{i=1}^N \phi (y(t_i)) \, .
\end{equation}
\end{subequations}
We further note that 
{\rev 
\begin{equation}\label{eq:phi2ker}
    \|\bar{\Phi}\|_F^2 = \tr(K^yH) \, , \qquad \|\bar{\Psi}(\omega)\|_F^2 = \tr(K^x(\omega)H) \, , \qquad H_{ij} :\,= \delta_{ij}-\frac{1}{N} \, ,
\end{equation}
where the Gram matrix $K^y$ is defined by
\begin{equation}\label{eq:gram}
     K^y _{ij} :\, = k^y(y(t_i), y(t_j)) \, , \qquad k^y(y,y'):\, = \langle \phi(y), \phi(y')\rangle \, ,
\end{equation}
for every $1\leq i,j \leq N$, and $K^x$ and $k^x$ are defined analogously to \eqref{eq:gram}; see the details in Appendix \ref{ap:phi2ker}.
}

We can therefore rewrite \eqref{eq:ml_nl_raw_a} as
\begin{equation}\label{eq:omega_mid}
\hat{\uomega}  = \arg \max_{\uomega \in \Omega} \frac{\langle \bar{\Psi}(\omega), \bar{\Phi}\rangle_F^2}{\|K^x(\omega)H\|_F^2 \|K^y(\omega)H\|_F^2} \, .
\end{equation}
In practice, we do not know what the the maps $\phi$ and $\psi$ are, and consequently cannot compute their corresponding kernels and Gram matrices \eqref{eq:gram}. However, the kernels $k^x$ and $k^y$ are both Mercer kernels, i.e., their Gram matrices~\eqref{eq:gram} are always positive semi-definite \cite{saitohBook}, they define an inner product on some (infinite-dimensional) reproducing kernel Hilbert Space $\mathcal{H}$.\footnote{The kernels $k^x$ and $k^y$ are generally {\em not} inner products in the original spaces $\mathbb{R}^d$ and $\mathbb{R}^D$, respectively, since they are not linear in either of their components.} This observation suggests that one should apply a well known heuristic known as the ``kernel trick", where rather than estimating the feature map $\psi$ from an infinite-dimensional function space, one chooses the kernels $k^x$ and $k^y$. The choice of kernels reflects our understanding and prior knowledge of the data and what makes two samples $x(t_i)$ and $x(t_j)$ similar (and similarly for $y$). The kernel trick here seems necessary, since we do not know the observation function~$G$. If we were to reconstruct $G$, then we could have used $\psi = G$, $\phi = {\rm Id}$, and $\mathcal{H}= \mathbb{R}^D$, see e.g. \cite{boots2012two}. In our setting, however, recovering $G$ might be nearly impossible, since we have only a single output time-series $y(t)$ and do not know the parameter vector $\omega^*$.

We revisit \eqref{eq:omega_mid} with the kernel trick in mind. The numerator $\langle \bar{\Psi}(\omega), \bar{\Phi}\rangle_F^2$ is equal to $\text{Tr}^2(C_{xy}(\omega))$, where $C_{xy}(\omega)= \bar{\Psi}(\omega)\bar{\Phi}^{\rm T}$ is the cross-covariance operator. This operator norm cannot be computed, since we only choose the kernels $k^x$ and $k^y$ and not the feature maps $\phi$ and $\psi$. Nonetheless, we can use the kernels $k^x$ and $k^y$ to estimate ${\rm Tr}(C_{xy}^{\rm T} C_{xy} ) = \|C_{xy}(\omega)\|_F^2$ as a surrogate to ${\rm Tr}^2(C_{xy})$. The norm~$\|C_{xy}(\omega)\|_F^2$ is known as the Hilbert Schmidt independence criterion (HSIC), due to Gretton et al. \cite{gretton2005measure, gretton2005kernel}, and can be estimated empirically by ${\rm Tr}(K^x(\omega)HK^yH)$. We therefore use the ${\rm HSIC}(\omega)$ to define the following realizable proxy of objective \eqref{eq:ml_nl_raw}:

\begin{equation}\label{eq:w_ml}
\hat{\uomega}  = \arg \max_{\uomega \in \Omega} ~ \frac{{\rm HSIC}(\omega) }{\|\bar{\Psi}(\omega)\|_F^2\cdot \|\bar{\Phi}\|_F^2 } = \arg \max_{\uomega \in \Omega} \frac{{\rm Tr}(K^x(\omega)HK^yH)}{\|K^x(\omega)H\|_F^2 \|K^y(\omega)H\|_F^2} \, .
\end{equation}

 The HSIC is an indicator for the dependence of $\phi$ and $\psi$ (or $x$ and $y$) as random variables. As $y(t)$ is determined by $x(t;\omega^*)$ up to a noise term, we expect that ${\rm HSIC}(\omega^*)$ would express a high statistical dependency. However, HSIC is also maximized as the covariance-matrix of $x$ (or of $\Psi$ in the nonlinear settings) is maximized, regardless of $C_{xy}$. We therefore normalize the HSIC estimator by the norm of the standard deviation matrix $\|\left(\bar{\Psi}(\omega)\bar{\Psi}(\omega)^{\rm T}\right)^{1/2}\|_F^2$. This latter matrix is estimated in the nonlinear Hilbert space settings by $\|K^x(\omega)\|_F^2$, which leads to the right-hand side of \eqref{eq:w_ml}. The empirical estimator of HSIC in the nominator on the right side is known to have a bias of ${\mathcal{O}}(1/N)$ \cite{gretton2005measure}.

{\bf Different perspective:} The estimator $\omega_{\rm ml}$ in \eqref{eq:w_ml} can also be viewed in terms of kernel density estimators (KDE). In \cite{vestner2017product}, the authors study non-rigid shape correspondence in three-dimensional objects. Given $N$ points $x_i$ and $y_i$ on two respective deformations of the same shape, the goal is to find a permutation $\pi$ on $1,\ldots N$ such that each $y_i$ corresponds to $ x_{\pi (i)}$ in the deformed shape. Letting~$K^x(\pi)$ be the Gram matrix of the permutated $x_i$'s, the term ${\rm Tr}(K^x(\pi)K^y)$ can be understood as the KDE estimator of the joint probability of the points under the permutation $\pi$. Indeed, in these settings the maximum likelihood estimator of $\pi$ among all permutations maximizes ${\rm Tr}(K^x(\pi)K^y)$, much as in Section \ref{sec:analysis} of this paper. Note that since the parameter estimated in \cite{vestner2017product} is a permutation, $\|K^x(\pi)\|_F^2$ is constant (independent of $\pi$) and therefore the normalization is not needed.

In \cite{michaeli2016cca}, the authors use the kernel trick to solve the nonparametric CCA problem, i.e., identify nonlinear mappings $\psi$ and $\phi$ that maximize the correlation between $\psi(x)$ and  $\phi(y)$ (as in the numerator of \eqref{eq:ml_nl_raw_a}). They show that the solution can be expressed using the singular values of the joint probability density of $X$ and $Y$, and use kernels to estimate this joint density in a nonparametric fashion. In the context of our problem, this solution can be used after $\omega^*$ is estimated for comparing the trajectories in a low dimensional representation.

\section{Proposed method}
\label{sec:method}

\subsection{Exhaustive search approach} In light of the analysis of Sec.\ \ref{sec:analysis}, our first proposed method, Algorithm \ref{alg:search}, is a straightforward numerical application of \eqref{eq:w_ml}. Assume for simplicity that $\Omega$ is a box in $\mathbb{R}^m$, i.e., $\Omega = \prod_{i=1, \ldots , m} [a_i, b_i] $ where $b_i > a_i$ for all $i=1, \ldots , m$. {\rev 
We define the Gram matrices for $x$ and $y$ by 
\begin{equation}\label{eq:kernely}
K^{y}_{i,j} :\,= \exp\left(- \frac{\|y(t_i) - y(t_j) \|_2 ^2}{\varepsilon_y} \right) \, , \qquad \varepsilon_y > 0 \, .
\end{equation}

\begin{equation}\label{eq:kernelx}
K^{x}_{i,j}(\utheta) :\,= \exp\left(- \frac{\|x(t_i;\omega) - x(t_j;\omega) \|_2 ^2}{\varepsilon_x} \right) \, , \qquad \varepsilon_x > 0 \, .
\end{equation}
Then, following \eqref{eq:w_ml}, we propose the score
 \begin{equation}\label{eq:score}
s(\omega) = \frac{\trace(K^{x} (\omega) HK^{y}H)}{\|K^{x}(\omega)H\|_F \cdot \|K^{y}H\|_F}  \,  ,
\end{equation}
where $H_{ij}=\delta_{ij}-N^{-1}$, which Algorithm \ref{alg:search} maximizes on a pre-determined set of grid-points in $\Omega_{\rm search} \subset \Omega$.

}

As noted in Sec.\ \ref{sec:nl}, since we do not know the feature maps $\phi$ and $\psi$, one needs to choose the kernel $k^x$ and $k^y$ (or their Gram matrices $K^x$ and $K^y$) to use~\eqref{eq:w_ml}. There are many possible choices of kernels which reflect different notions of affinities between samples of $x(t)$ (and of $y$), many of which might have worked well for estimating $\omega^*$, see e.g., \cite{hofmann2008kernel}. In this work we choose the widely popular Gaussian kernel $k^x (x(t_i),x(t_j))=\exp(-\|x(t_i)-x(t_j)\|^2/\varepsilon_x )$ (and respectively for $y$), for two main reasons: first, the Gaussian kernel is translation invariant, i.e., $k^x(x,x')=k(x-x')$, which ensures we capture only relative changes in the data, and not absolute values. Second, the exponential decay of the Gaussian kernel attenuates the effect of large distances. For the model coordinates $x(t)$, this makes intuitive sense because the governing ODE \eqref{eq:ode} is local. For the observation coordinates, attenuating large distances counteracts the spuriously large pairwise distances which tend to appear in $\ell^2(\mathbb{R}^D)$ for $D\gg 1$, see e.g., \cite{high-d,high-d2}. {\rev Finally, the Gaussian kernel has expressivity properties that ensure its ability to capture polynomial-order nonlinear dependencies between different times, but it is by no means the only possible choice. For detailed analysis and alternative kernel choices, see  \cite{smola1998learning} and the references therein. A key takeaway from this work is that even though other tailored kernels can be designed for better performance, the standard choice of the Gaussian kernel yields good results in our numerical experiments.}

\begin{algorithm}[htbp]
\caption{Kernel search-based method for estimating $\omega ^*$}\label{alg:search}
Given $\{y(t_n)\}_{n=1,\ldots , N}$. 
\begin{algorithmic}[1]
\STATE{Compute the $y$ kernel \eqref{eq:kernely}.}
\STATE{Choose $\Omega_{\rm search} \subset \Omega$ to be a finite Cartesian grid in $\Omega \subseteq \mathbb{R}^n$.}
\FOR{ each $\omega \in \Omega_{\rm search}$}
\STATE{Solve \eqref{eq:ode} for $x(t;\omega)$.}
\STATE{Compute the $x$ kernel \eqref{eq:kernelx}.}
\STATE{Compute the score $s(\omega)$, see \eqref{eq:score}.}
\ENDFOR
\RETURN{ $\hat{\omega} =\arg \max\limits_{\omega \in \Omega_{\rm search}} s(\omega)$.}
\end{algorithmic}
\end{algorithm}

The effectiveness of Gaussian kernels strongly relies on proper tuning of the kernels' bandwidths $\varepsilon_x$ and $\varepsilon_y$. These parameters directly affect the feature maps induced by the kernels. At one extreme, setting $\varepsilon_x$ or $\varepsilon_y$ too large would result in kernels that approach the all-ones matrices, i.e., where all samples are equally affine. At the other extreme, as $\varepsilon_x\rightarrow 0$ (resp.\ $\varepsilon_y$), the kernel $K$ approaches the identity matrix, i.e., all affinities between different samples are neglected. Here, we use a max-min measure suggested in \cite{Keller} where the scale is set to
\begin{equation} \label{eq:MaxMin}
\varepsilon_{y}= \underset{j}{\max} [ \underset{i,i\neq j}{\min} (||y(t_i)-y(t_j)||^2)] \, , \qquad i,j=1,...N \, ,
\end{equation}
 and analogously for $\varepsilon_x$. The max-min approach guarantees that for each data-point, $K$ expresses a non-negligible affinity with at least one other point. Moreover, this scheme generates kernels $K^x$ and $K^y$ that are invariant to scaling in the ambient observation space $\mathbb{R}^D$. Several other methods have been proposed for tuning $\varepsilon$ and can certainly be used, see for example \cite{Keller,zelnik,epsilon,Singer}. The third tunable parameter in Algorithm \ref{alg:search} is the grid size $|\Omega_{\rm search}|$.  The choice of a predeteremined grid $\Omega_{\rm search} $ affects Algorithm \ref{alg:search} in two ways:
\begin{enumerate}
    \item {\bf Accuracy:} For any estimator $\hat{\omega}$ of $\omega^*$, define the estimation error
    \begin{equation}\label{eq:estimation_err}
{\rm Err}\, \omega _j =\frac{ \|\hat{\omega}_j-\omega_j^* \|_2}{\omega_j^*} \, , \qquad j=1,..,m \, .
\end{equation}  
\noindent
Generally in Algorithm \ref{alg:search}, the grid does not necessarily include $\omega^*$, i.e., $\omega^* \not\in~\Omega_{\rm search}$, and so $\hat{\omega}\neq \omega^*$. Therefore, even in the best case scenario where Algorithm~\ref{alg:search} returns the closest grid point to $\omega^*$, the error is typically bounded from below in terms of $\Delta \omega$. {\rev This error estimation applies also to the case of a single parameter, as the average estimation error~\eqref{eq:estimation_err} over many experiments scales like  $\Delta\omega$, where $$\Delta \omega:\,= \min \{|\omega_i - \omega_j | ~~ {\rm s.t.}~ \omega_i, \omega_j \in \Omega_{\rm search} \, , ~ i\neq j \} \, ,$$ is the spacing of the grid. To see that, suppose for simplicity that $\Omega = [0,\Delta \omega]$ and that $\omega^*$ is drawn uniformly at random from $\Omega$, i.e., the probability density function of $\omega^*=y$ is $p(y)=\Delta \omega ^{-1}$. Then by partitioning $\Omega$ to two halves, one closer to the grid point $\omega = 0$ and the other to the grid point $\omega = \Delta \omega$, we have that 
$$\mathbb{E}_{\omega^*} \, {\rm dist}(\omega^*, {\rm grid}) =  \frac{1}{\Delta \omega} \left( \int\limits_{0}^{\Delta \omega/2} \omega^* \, d\omega^* + \int\limits_{\Delta \omega /2}^{\Delta \omega} (\Delta\omega - \omega ^*) \, d\omega^* \right) = \frac{\Delta \omega}{4} \, .  $$
A similar estimate holds for any dimension $m\geq 1$ of $\Omega$.}
\item {\bf Efficiency:} In the multi-dimensional case $\Omega \subseteq \mathbb{R}^m$, fine grids are computationally prohibitive since their size scales exponentially with $m$. Large grids are especially an issue when either {\it (i)} solving the underlying dynamical system \eqref{eq:ode} is computationally expensive, {\it (ii)} the length of the time series $N$ is large. In the latter case, the computation of the kernel \eqref{eq:kernelx}, which requires evaluating all pairwise distances, requires $\mathcal{O}(N^2)$ operations. The cost of the kernel computation can be reduced using methods such as $k$-sparse graph \cite{FastKNN} which enjoys a reduced complexity of ${\mathcal{O}}(N log N + Nk)$, where $k$ is the number of nearest neighbors used for building the graph. 
\end{enumerate}

\subsection{Optimization approach}

To overcome both of the accuracy and the efficiency problems of Algorithm \ref{alg:search}, we would like to replace the exhaustive grid search with an optimization scheme to solve the following problem
\begin{equation}\label{eq:opt_prob}
\begin{array}{l}
 \text{maximize } s(\omega) = \frac{\trace(K^{x} (\omega) HK^{y}H)}{\|K^{x}H(\omega)\|_F \cdot \|K^{y}H\|_F}  \\
 \text{over } \omega \in \mathbb{R}^m \\
 \text{subject to } R(\omega) \leq 0 \, , 
 \end{array}
 \end{equation}
where $R$ can be chosen to be any function such that $R(\omega)\leq 0$ if and only if $\omega \in \Omega$.\footnote{It can often be the case that for $\omega$ values outside of $\Omega$, the underlying ODE does not yield a well-posed solution. For example, the harmonic oscillator $\ddot{x}(t) + \omega x = 0$ for $\omega<0$ yields exponentially growing and ill-posed solutions.} Critically, the optimization problem \eqref{eq:opt_prob} is in general non-convex. This is an {\em inherent feature of our problem} and should not depend on the specific solution strategy. Indeed, let~$\tilde{s}(\omega)$ be any convex cost function for which $\omega^* = \arg \max \tilde{s}(\omega)$. Since $\tilde{s}$ depends on $\omega$ indirectly through $f(x;\omega)$ (see \eqref{eq:ode}) and since there is no unique way to express the dependence of $f$ on its parameters, one can find an equivalent parametrization of the ODE \eqref{eq:ode} such that the resulting $\tilde{s}(\omega)$ is no longer convex. Therefore, the standard form of our ODE of interest need not result in a convex parametrization of~$\tilde{s}(\omega)$. We note that this non-convexity property is already true for the linear model~\eqref{eq:ml_ip}.

\begin{algorithm}[htbp]
\caption{Kernel optimization-based method for estimating $\omega ^*$}\label{alg:opt}
Given $\{y(t_n)\}_{n=1,\ldots , N}$. 
\begin{algorithmic}[1]
\STATE{Compute the kernel for $y$ \eqref{eq:kernely}.}
\STATE{Choose $\Omega_{\rm init} \subset \Omega$ at random.}
\FOR{ each $\omega_j \in \Omega_{\rm init}$}
\STATE{Solve the optimization problem \eqref{eq:opt_prob} to find $\hat{\omega}_j$ using interior point optimization initialized at $\omega_j$, where the kernel $K^x (\omega)$ is given by \eqref{eq:kernelx} and $x(t;\omega)$ are the solutions of \eqref{eq:ode}. }
\ENDFOR
\RETURN{ $\hat{\omega} =\arg \max\limits_{j} s(\hat{\omega}_j)$.}
\end{algorithmic}
\end{algorithm}

To solve the constrained, nonlinear and non-convex problem \eqref{eq:opt_prob}, our proposed method, Algorithm \ref{alg:opt}, has ``two layers" of optimization. At the heart of Algorithm~\ref{alg:opt}, we use the Interior-point Algorithm~(IPA), a solver for nonlinear constrained optimization problems \cite{IPA1,IPA2}, as the optimization scheme in Step 4 of Algorithm~\ref{alg:opt}.\footnote{In our simulations, we used {\sc MATLAB}'s implementation of the IPA method.} The IPA method first takes a \textit{Newton step} by attempting to solve a linear approximation of the problem \eqref{eq:opt_prob}, then, a gradient step is performed using a trust region \cite{IPA3}. Since the problem is not convex (see details in the next paragraph), we repeatedly initialize the IPA method at random points $\Omega_{\rm init}\subset \Omega$, and then chooses the optimal result over all iterations. In our experience, the multiple initialization mechanism improves our chances of finding the global maximum; see experimental results in Sec.\ \ref{sec:results}.

Since the IPA routine in Algorithm \ref{alg:opt} dynamically samples $\omega \in \Omega$ values, the overall number of samples does not scale exponentially with the dimension, and the accuracy is not limited by the grid spacing. Algorithm \ref{alg:opt} is therefore computationally cheaper than Algorithm \ref{alg:search}. The key parameter in comparing the two is the number of times
the ODE \eqref{eq:ode} is solved and~$K^x$~ \eqref{eq:kernelx}~is computed. In Algorithm~\ref{alg:search}, this is exactly the grid size $|\Omega _{\rm search}|$. In Algorithm~\ref{alg:opt}, the number of evaluations of~$x(t;\omega)$~is the number of optimization initializations~$|\Omega_{\rm init}|$ multiplied by the number of iterations in each optimization process. In the examples considered in this study, $|\Omega _{\rm init}|$ was kept orders of magnitude smaller than the $|\Omega_{\rm search}|$ without much loss of accuracy, and the number of iterations in each optimization process is usually below $20$; see e.g., Fig.\ \ref{fig:GD}. Algorithm \ref{alg:opt} therefore suggests an avenue to solve~\eqref{eq:opt_prob} efficiently and accurately even as the dimension $m$ of the parameter space~$\Omega$~increases.

\begin{remark}
Another practical implementation aspect of Algorithms \ref{alg:search} and \ref{alg:opt} is the numerical solution method of the ODE \eqref{eq:ode}. Throughout this paper we used the standard fourth order Runge-Kutta method \cite{iserles2009numerical}. Other numerical methods for ODEs can be used; see \cite{keller2020discovery} for a more thorough discussion. 

\end{remark}

\subsection{Degeneracies and Identifiability}\label{sec:degen}


When is the problem of estimating $\omega^*$ unsolvable? If $x(t;\omega)=x(t;\omega ^*)$ for some $\omega  \neq \omega ^*$, these two parameters are indistinguishable in terms of the resulting dynamics. Moreover, if $G(x(t;\omega)) = G(x(t;\omega ^*))$, then the experiment/observation of the dynamical system cannot distinguish between the parameters.  {\rev These are extreme cases for obstructions of identifiability and observability, topics which have been studied in both the statistics and control literature, see e.g., \cite{nguyen1982ident, villaverde2019obs} and the references therein. The system \eqref{eq:ode} combined with the observation function \eqref{eq:Gdef} is said to be unidentifiable if the problem of estimating $\omega^*$ is not solvable, regardless of the method of solution.  In loose terms, such $G$ then corresponds to a flawed experiment which is not designed to estimate $\omega$. 

Even for an identifiable and observable system, the fact that the observation function $G$ is unknown limits our use of standard inference techniques. In what follows we wish to discuss the limitation of our method even for identifiable systems:} suppose that neither $x$ nor $G$ are degenerate (as functions of $\omega$ and $x$, respectively), but that the Gaussian kernels in \eqref{eq:kernely} and \eqref{eq:kernelx} are degenerate. To explore the effect of these degeneracies, we will consider the case where $G$ is an $\ell^2(\mathbb{R}^d\to \mathbb{R}^D)$ isometry and noiseless, i.e., $\|G(x_1)-G(x_2)\|_2 = \|x_1-x_2\|_2$ for any $x_1, x_2\in \mathbb{R}^d$. This discussion highlights some of the considerations that go into designing $K^x$ and~$K^y$.

\begin{lemma}\label{lem:degeneracy}
Let $y(t)=G(t)$ where $G$ is an $\ell^2(\mathbb{R}^d \to \mathbb{R}^D)$ isometry. Then $$s(\omega^*)=1=\max_{\omega \in \Omega} s(\omega) \, .$$
The other $\omega\in \Omega$ values where $s(\omega)=1$ are precisely those for which $x(\omega)=Tx(\omega^*)$ for some $\ell^2(\mathbb{R^d})$ isometry $T$.
\end{lemma}
See the proof in Appendix \ref{ap:normpf}. Intuitively, Lemma \ref{lem:degeneracy} implies that while $s(\omega)$ is not uniquely maximized in this case, it is ``reasonably degenerate", in the sense that its only other maximizers (when $G(x)=x$) are $\omega \in \Omega$ for which the trajectory $x(t;\omega)$ is isometric to $x(t;\omega^*)$. For example, if $x$ is a scalar, it would mean that such degeneracies are only translations and reflections.

This is a fundamental consideration in the kernel design - if $\omega, \omega^* \in \Omega$ manifest in isometric observations/trajectories, then one cannot distinguish between them using the observations. The assumption that underlies our choice of the $\ell ^2$ norm in the kernel~\eqref{eq:kernelx} now becomes apparent - it expresses the kind of degeneracies we wish to allow. Indeed, different choices of norms would yield different equivalence classes of parameters in $\Omega$.

Finally, we make a crucial note regarding our choice of the Gaussian kernel. In the proof of Lemma~\ref{lem:degeneracy} we show that, in ideal settings, $s(\omega)$ is maximized only when $\Delta = K^x(\omega)-K^x(\omega^*)=0$. However, since the kernel \eqref{eq:kernelx} is {\em exponentially decaying} with $\|x_i - x_j\|_2^2$, the entries $\Delta_{ij}$ are practically null for most far-away indices (times) $i$ and $j$. Hence, even local in time degeneracies are sufficient to cause  estimation error. As noted above, the choice of $\varepsilon_x$ and $\varepsilon_y$ determines how ``local" the respective kernels are.

\section{Experimental Results}\label{sec:results}
 To test Algorithms \ref{alg:search} and \ref{alg:opt}, we apply them to two classical chaotic dynamical systems - the double pendulum and the Lorenz system.

\subsection{Double Pendulum}

\label{sec:pend}
The double pendulum consists of two pendulums, one attached to the end of the other. The Lagrangian of this system is given by \cite{landau1980mechanics}
$$ L= \frac12 \left(m_1+ m_2\right)l_1 ^2 \dot{\theta}_1 ^2 +\frac12 m_2l _2 ^2 \dot{\theta}_2 ^2 +m_2l_1 l_2 \dot{\theta}_1\dot{\theta}_2 \cos(\theta_1 - \theta_2) \, ,$$
where $l _j, m_j$, and $\theta_j(t)$ are the length, mass, and clock-wise angle from the negative~$y$ direction of the $j$-th pendulum, for $j=1,2$. From this Lagrangian, a fourth-order system of Euler-Lagrange ODEs for $(\theta_1, \theta_2, \dot{\theta}_1, \dot{\theta}_2)$ can be derived; see e.g., \cite{shinbrot1992chaos} and Appendix \ref{ap:double}.

\begin{figure}[htb!]
\begin{center} 
\includegraphics[width=0.45\textwidth]{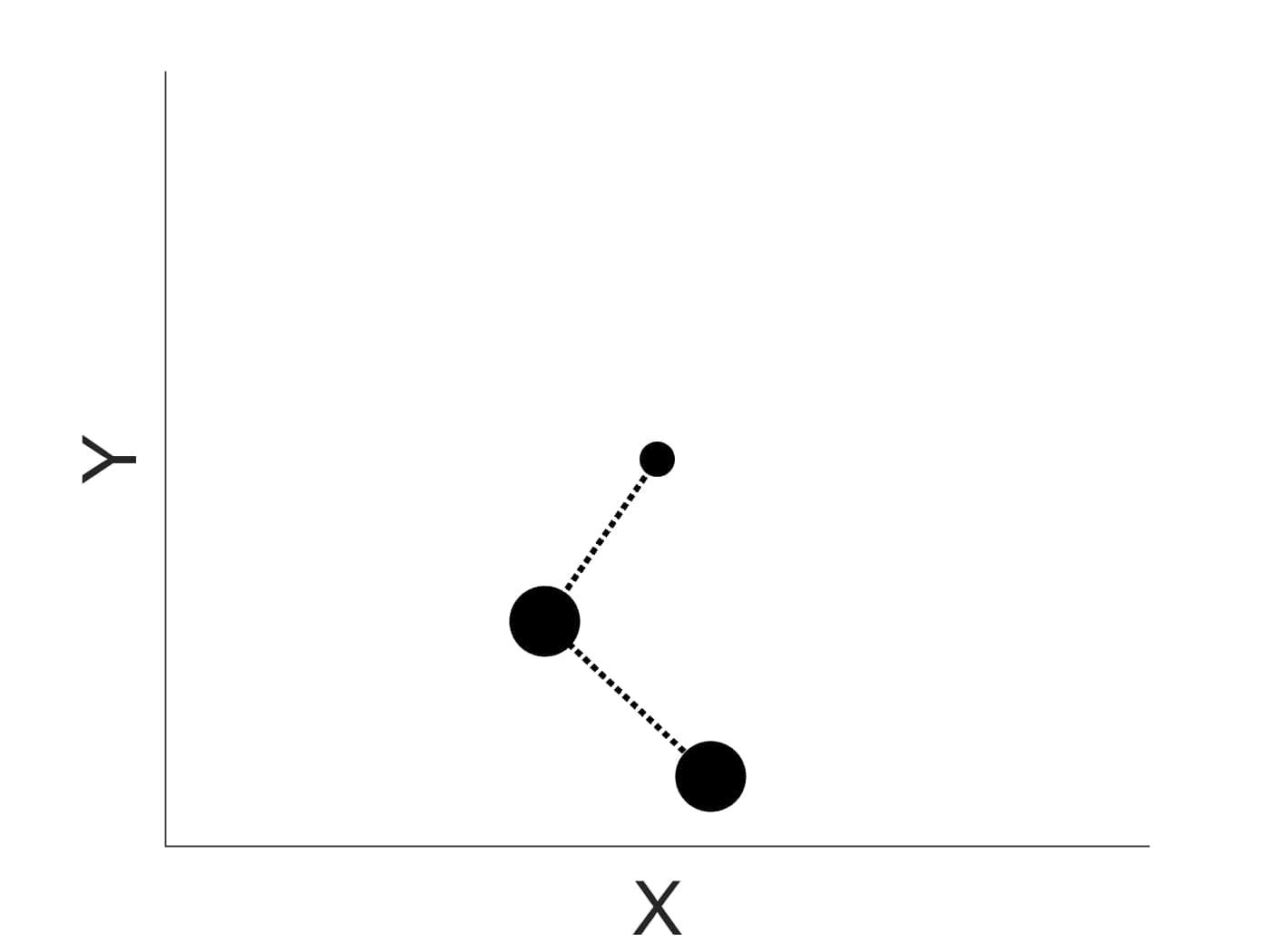}
\includegraphics[width=.45\textwidth]{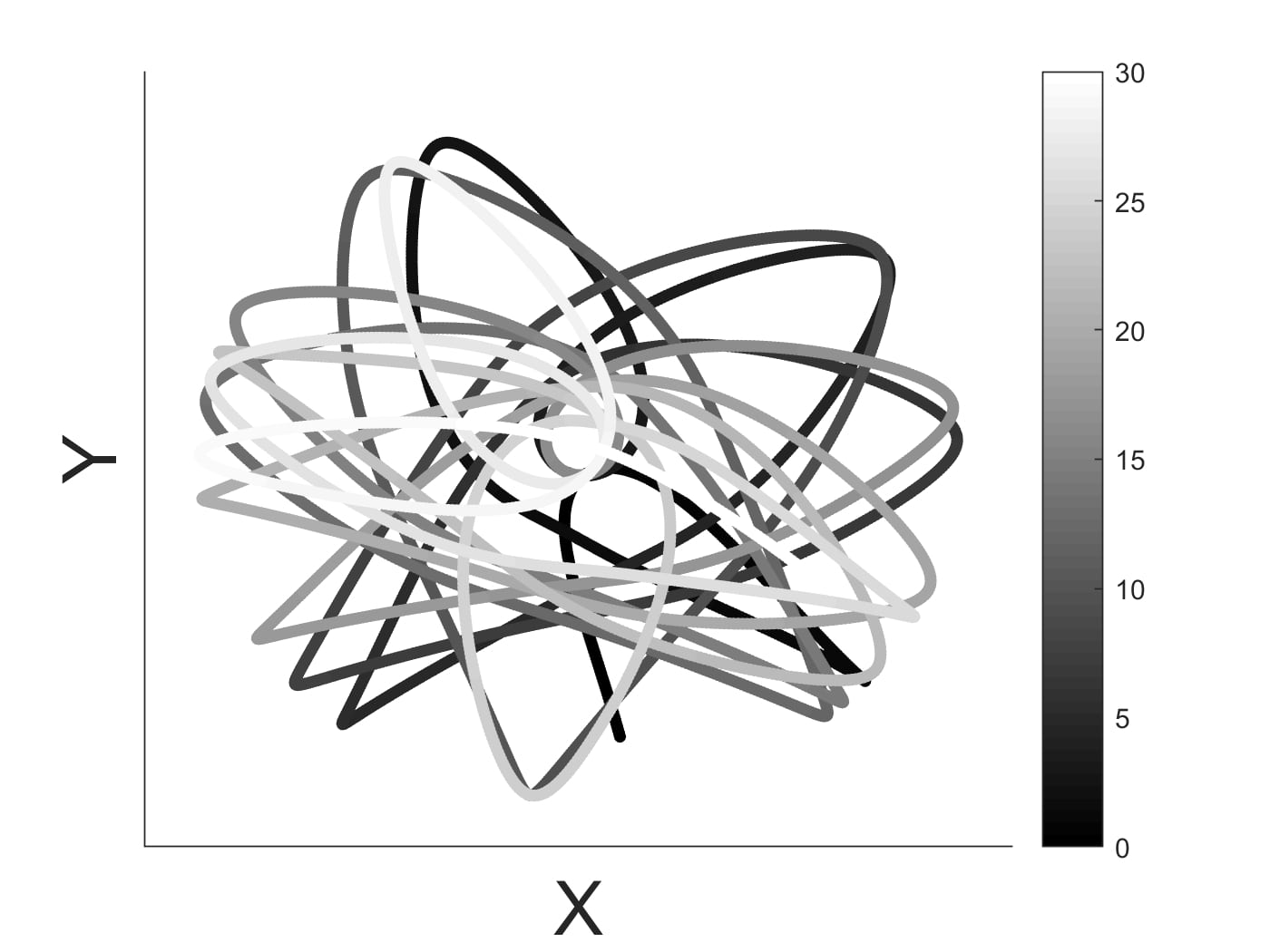}

\end{center}
\caption{Left: A frame of an artificial double pendulum video. Right: The chaotic trajectory $(x_2(t), y_2(t))$ of the bottom bob.}
\label{fig:pend_example}
\end{figure}

\begin{figure}[htb!]
\begin{center} 
\includegraphics[width=1\textwidth]{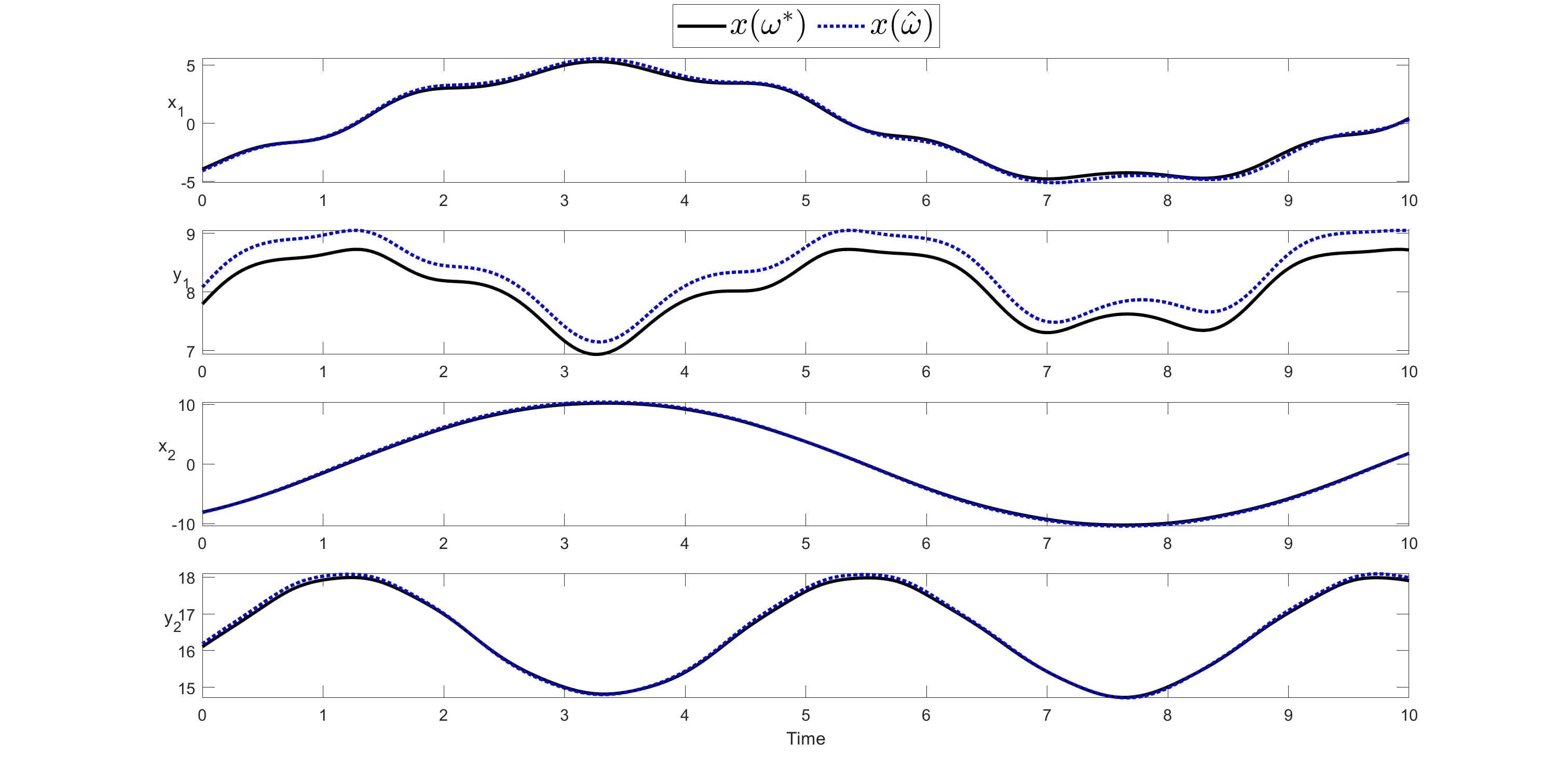}
\end{center}
\caption{Application of Algorithms \ref{alg:search} to estimate the double pendulum parameters $\omega = (l_1, l_2, m_2)$. The horizontal and vertical coordinates $x_i$ and $y_i$ of the two bobs ($i=1,2)$ for the true parameter $\omega^*$ (black) and using $\hat{\omega}$ when estimated using Algorithm~\ref{alg:search}~(dashed blue). }
\label{fig:pend_example_res}
\end{figure}

The parameters of the system are $\uomega=(m_2,l_1,l_2)$, where we set $m_1=1$ since the motion of the double pendulum depends only on the ratio  $m_2/m_1$; see Appendix~\ref{ap:double}. We describe the dynamics of the double pendulum using the Cartesian coordinates of the two bobs $x=(x_1,y_1,x_2,y_2)$ (with a slight abuse of notations). The dynamics of the pendulum are observed through a synthetic video with a frame rate of~$\Delta t=~0.01$. Each frame embeds the model in a high dimensional space $y(t_i)\in \mathbb{R}^D$, where $D=~171 \times 217 =37,107$ is the number of pixels in each frame; see Fig. \ref{fig:pend_example}.

\begin{figure}[htb!]
\begin{center} 

\includegraphics[width=0.45\textwidth]{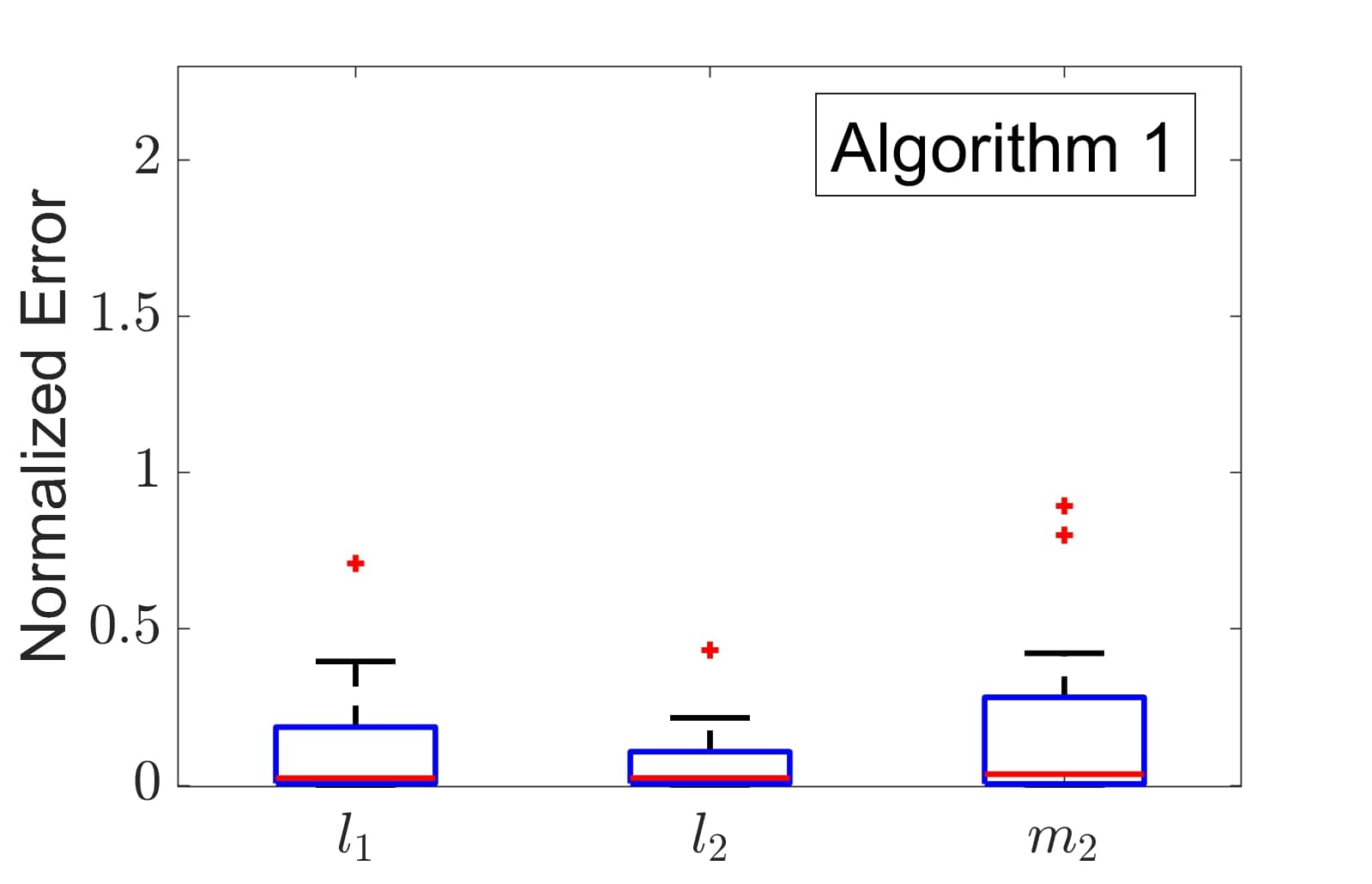}
\includegraphics[width=0.45\textwidth]{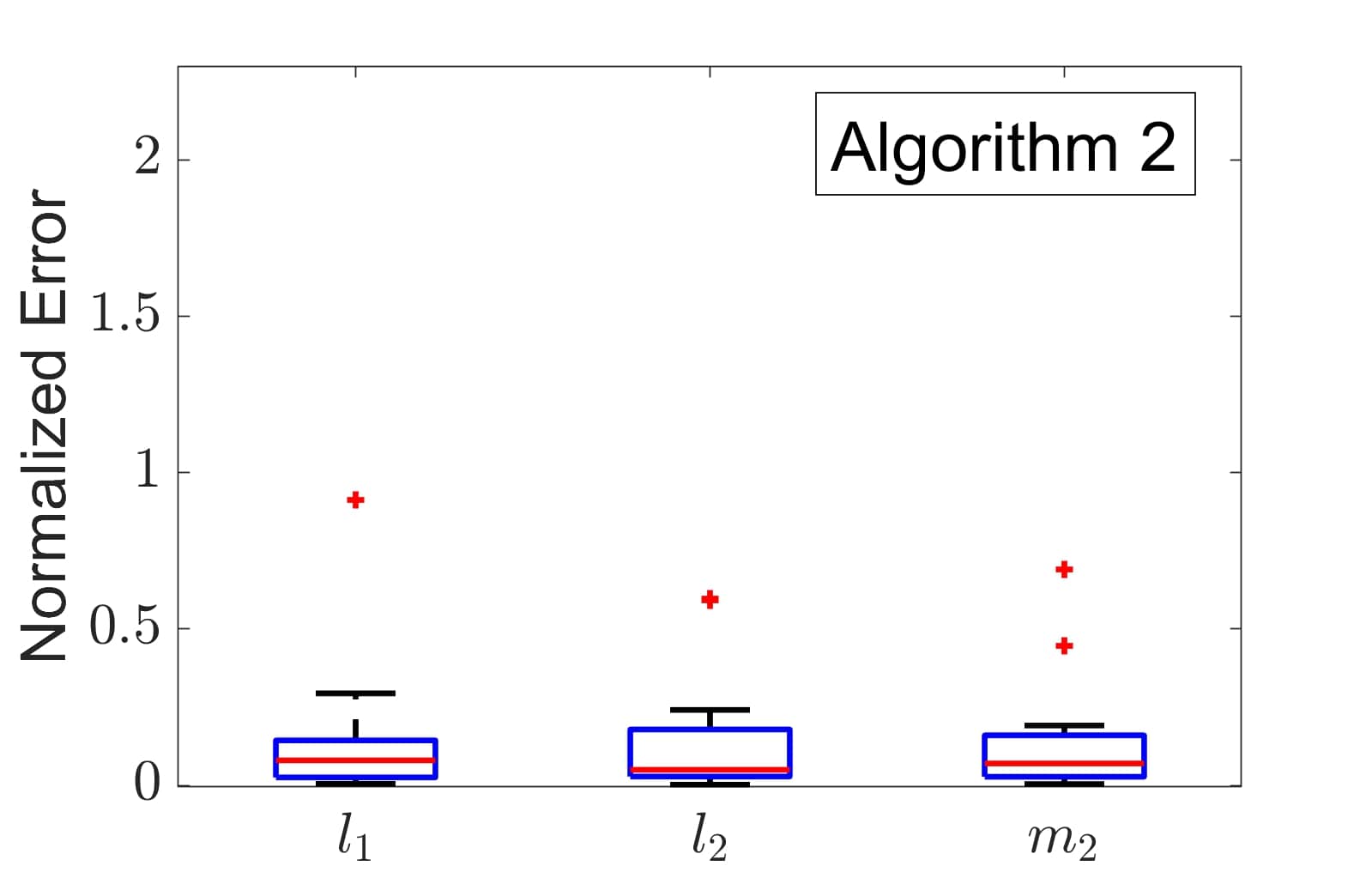}
\includegraphics[width=0.45\textwidth]{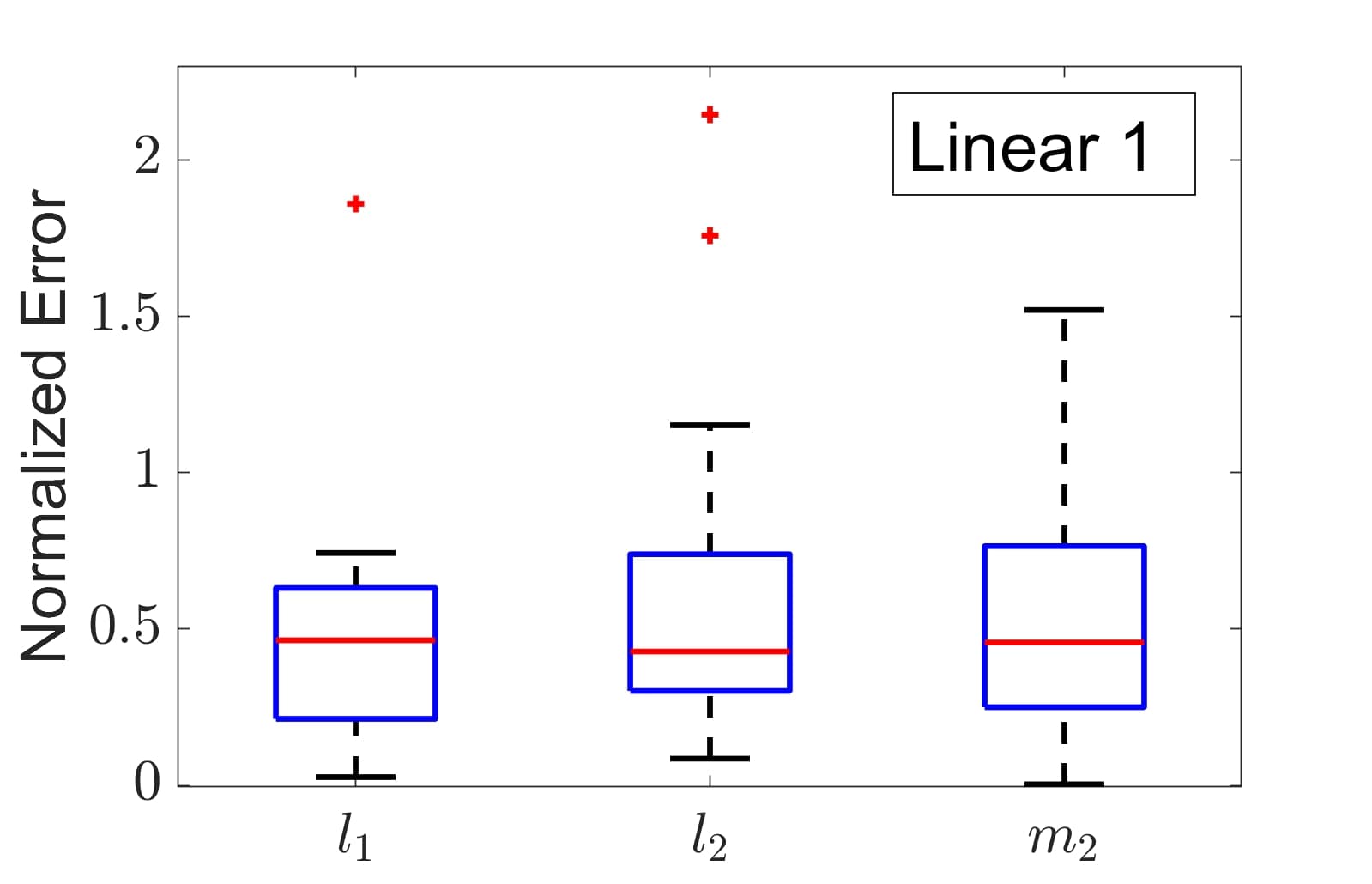}
\includegraphics[width=0.45\textwidth]{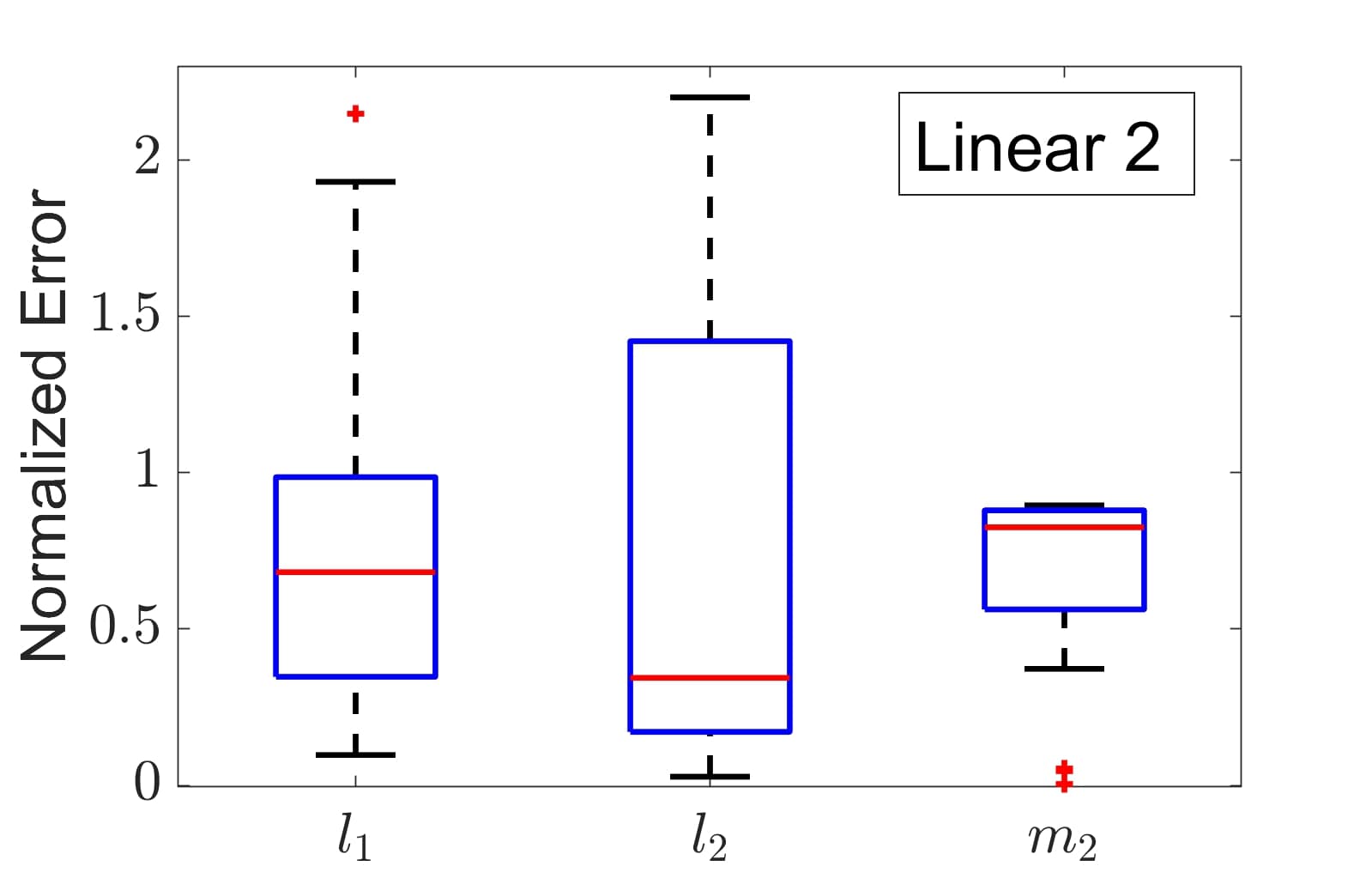}
\end{center}
\caption{Application of Algorithms \ref{alg:search} and \ref{alg:opt} to estimate the double pendulum parameters $\omega = (l_1, l_2, m_2)$. Top: Box plots of normalized errors \eqref{eq:estimation_err} of estimated parameters $l_1,l_2$ and $m_2$ for the double pendulum based on $20$ test cases. Results base on Algorithm \ref{alg:search} (left) and Algorithm \ref{alg:opt} (right). Bottom: Box plots of normalized errors for the linear algorithms 1 and 2, \eqref{eq:lin1} and \eqref{eq:lin2}, respectively. In all box plots, red lines represent the medians, the blue box represents the $25$-th and $75$-th quantiles, the black whiskers represent $\pm 2.7$ standard deviations, and any data point beyond this distance is considered as an outlier and marked by a red plus.}
\label{fig:pend_box_plots}
\end{figure}

To evaluate the proposed approach, we generate $20$ instances of a double pendulum with the parameters $m_2,l_1,l_2$ each drawn iid from a uniform distribution in the interval $[1,10]$ and with initial conditions randomly drawn independently from $\mathcal{N}(0,0.5)$, where $N=1,000$. We then generate from each instance a movie of total run-time $T=N \cdot \Delta t=10$. First, we apply Algorithm \ref{alg:search} to each movie and search over $20$ values of each parameter (where the initial conditions are known), i.e., a search-grid of size $| \Omega_{search} |=20^3=8,000$ and $\Delta \omega =0.45$. In Fig.\ \ref{fig:pend_example_res}, we see that the estimated parameters yield pendulum trajectories nearly indistinguishable from the true ones~(Fig.\ \ref{fig:pend_example_res}). Overall, the median normalized parameter estimation error \eqref{eq:estimation_err} of Algorithm \ref{alg:search} is $7 \%$, $8 \%$, and $5 \%$ for $m_2$, $l_1 $, and $l_2$, respectively. We repeat the same experiment for the optimization-based Algorithm \ref{alg:opt} with $|\Omega_{\rm init}|=1,000$, where the overall median normalized errors are $3.6 \%$, $2.3 \%$, and $2.4 \%$ for $l_1 $, $l_2$, and~$m_2$, respectively; see box-plots in Fig.\ \ref{fig:pend_box_plots}.  {\rev In this experiment, as well as in the Lorenz system (Sec. \ref{sec:lorenz}), we assume that the initial conditions are known. This is a realistic assumption if the experiment is controlled by the observer, or if the initial conditions can be estimated independently. If the initial conditions are not known, one option is to estimate them as additional parameters using Algorithm \ref{alg:search} or \ref{alg:opt}.}

To provide baselines for the proposed algorithms we use two linear variants of Algorithm \ref{alg:search}. Both variants are based on the same grid search procedure as in Algorithm \ref{alg:search} but using a different score. The score for the first linear estimator (Linear 1) is defined by 
\begin{equation}\label{eq:lin1}
 s^{\rm lin_1}(\omega) =\frac{\|\bar X \bar Y ^T\|_F}{\| \bar X\|_F\| \bar Y\|_F },  
\end{equation} where $\bar X$ and $ \bar Y$ are the centered versions of $X$ and $Y$. For the second linear estimator, we use Gram matrices instead of Gaussian kernels for $k^y$ and $k^x$ , thus the score is defined by
\begin{equation}\label{eq:lin2}
  s^{\rm lin_2}(\omega)= \frac{\trace(G^{x} (\omega) HG^{y}H)}{\|G^{x}H(\omega)\|_F \cdot \|G^{y}H\|_F} , 
\end{equation}
where $G^x=X^TX$ and $G^y=Y^TY$. As evident from the box-plots of Linear~1~and~2~(Fig.~\ref{fig:pend_box_plots}), the performance of the linear variants of Algorithm \ref{alg:search} are inferior compared with their kernel based counterparts. Specifically, the overall median error of Linear~1 is higher than Algorithm \ref{alg:search} by a factor of $6.5$. For Linear~2 this factor is $9.7$.

\begin{figure}[htb!]
\begin{center} 
\includegraphics[width=0.8\textwidth]{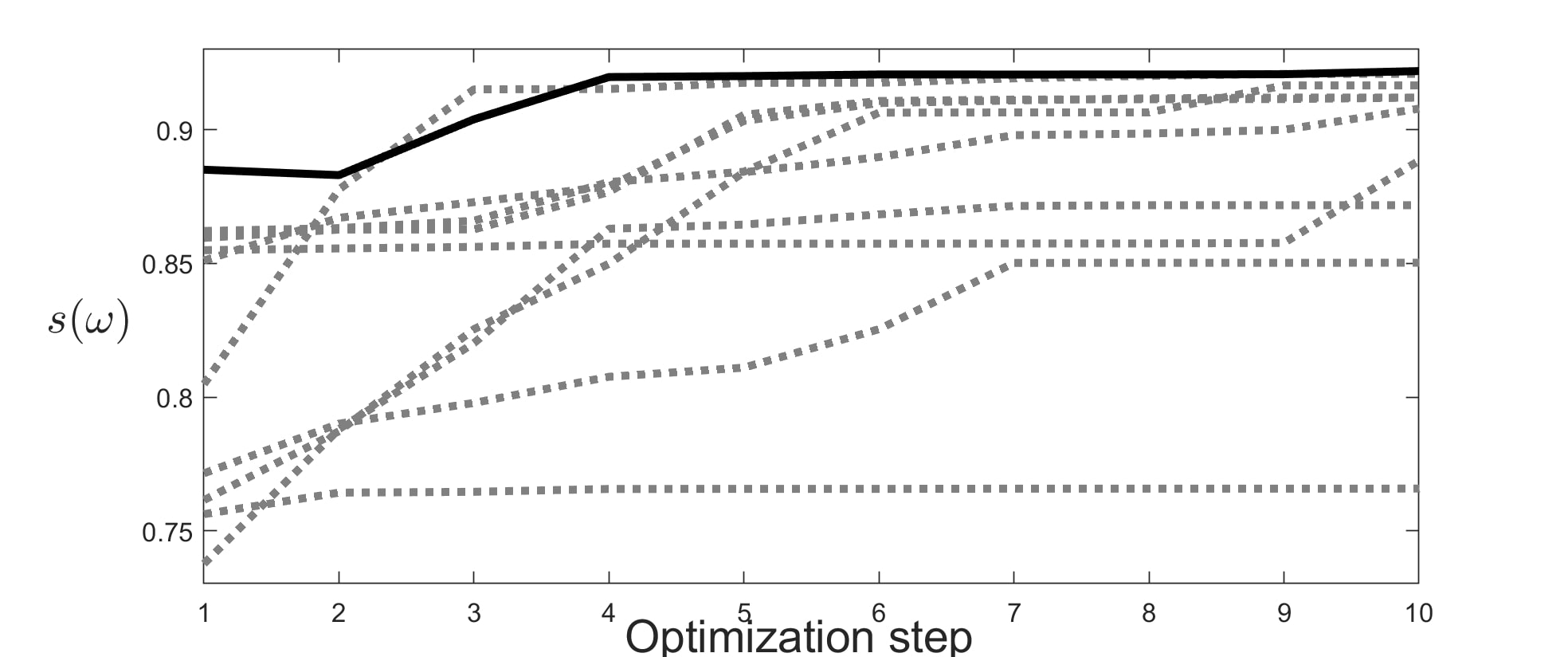}
\end{center}
\caption{$10$ runs of the IPA optimization method in Algorithm~\ref{alg:opt} for the case of the double pendulum; see Sec. \ref{sec:pend}. The score \eqref{eq:score} vs. the optimization step. Here the initial parameters for the IPA are drawn uniformly at random from $[0.9,1.1]\cdot\omega^*$. The solid line indicates the best IPA run in terms of the score $s(\omega)$.}
\label{fig:GD}
\end{figure}

Next, to demonstrate the efficacy of the optimization-based Algorithm \ref{alg:opt}, we record the values attained by the IPA method throughout its iterations. As can be seen in Fig.\ \ref{fig:GD}, most of the IPA runs converge after 10 iterations, and all of them converge in less than 30 iterations (results not shown). Some runs converge to parameters with a relatively low score. Nevertheless, since we choose the IPA run of the highest score, we find that the overall error of Algorithm \ref{alg:opt} is low.

In light of the discussion in Sec.\ \ref{sec:degen}, it is worth asking whether measuring the estimation error \eqref{eq:estimation_err} is the right way to estimate one's performance in estimating a dynamical system's parameters. Certainly, it might be that the problem is inherently ill-posed and that $\omega^*$ is non-identifiable if another parameter $\omega\in \Omega$ produce observations similar to $y(t)$. Along this reasoning, we propose another metric to measure our performance, the {\em prediction error}. This metric compares the normalized mean square error (MSE) of the predicted trajectory $x(t;\hat{\omega})$ from the true one $x(t;\omega^*)$
\begin{equation}\label{eq:prediction_err}
\frac{\int\limits_{t=0}^{t_{\rm f}} \|x(\tau ;\hat{\omega}) - x(\tau;\omega^*) \|_2^2 \, d\tau }{\int\limits_{t=0}^{t_{\rm f}} \|x(\tau ;\hat{\omega})  \|_2^2 \,d\tau }.
\end{equation}

\begin{figure}[htb!]
\begin{center} 
\includegraphics[width=0.6\textwidth]{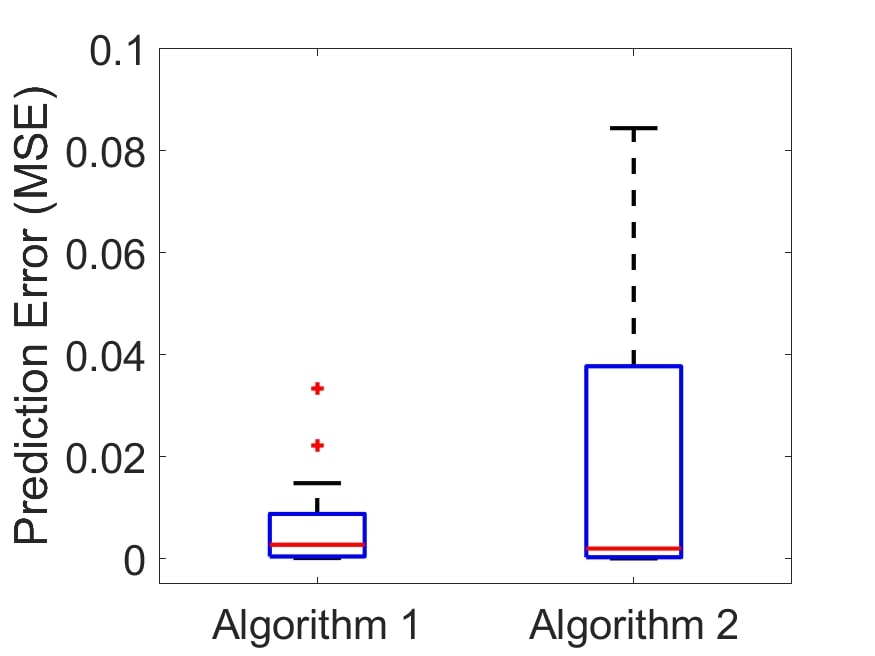}

\end{center}
\caption{Box plots of normalized prediction error \eqref{eq:prediction_err} for the double pendulum experiment, as in Fig.\ \ref{fig:pend_box_plots}, for Algorithms \ref{alg:search} and \ref{alg:opt}. }
\label{fig:mse_box}
\end{figure}

We evaluate the prediction error of Algorithms \ref{alg:search} and \ref{alg:opt} by comparing the true pendulum trajectory to the estimated trajectory using the same $20$ synthetic videos used for the box-plots in Fig.\ \ref{fig:pend_box_plots}. The results (see Fig.\ \ref{fig:mse_box}) demonstrate that the median prediction errors for Algorithm \ref{alg:search} and \ref{alg:opt} are $0.27\%$ and $0.2\%$ respectively. Moreover, both methods attain a prediction error smaller than $9\%$ in all of the simulations. In this experiment, the prediction error \eqref{eq:prediction_err} is comparable to the estimation error \eqref{eq:estimation_err} squared, as could be expected from their respective definitions.

\subsection{Estimation error decreases with signal length}\label{sec:length}

\begin{figure}[htb!]
\begin{center} 
\includegraphics[width=0.5\textwidth]{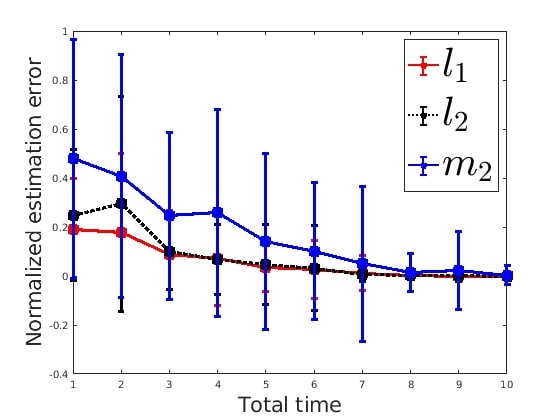}
\end{center}
\caption{Application of Algorithm \ref{alg:search} to the double pendulum systems with varying total run-time $T_{\rm f}$; see Sec.\ \ref{sec:length}. Mean normalized estimation error \eqref{eq:estimation_err} for the three parameters $l_1$, $l_2$ and $m_2$ as a function of the total length $T_{\rm f}$, with error-bars of one standard deviation.}
\label{fig:pend_tf}
\end{figure}

In many experiments, the overall run time $T_{\rm f} = N\cdot \Delta t$ of the experiment is a tunable parameter. How does~$T_{\rm f}$~affects the estimation error \eqref{eq:estimation_err}? In particular, we want to test the intuition according to which longer signals provide more information. For each $T_{\rm f}=1,2,\ldots, 10$ we draw 100 samples of $l_1,l_2,m_2$, {\rev each with the uniform distribution on the set $\{ 1.5, 2, 2.5, \ldots , 7\}$}. To each set of these parameters we apply Algorithm \ref{alg:search} as in the previous section. The mean estimation error \eqref{eq:estimation_err} decays as a function of $T_{\rm f}$, see Fig.\ \ref{fig:pend_tf}. Furthermore, it is evident that the standard deviation generally decreases with $T_{\rm f}$, which reinforces the intuition of more information in longer signals. 

We remark, however, that we do not expect the accuracy to increase with $T_{\rm f}$ for all dynamical systems, especially not for systems with attractors or limiting cycles. Consider for example $\dot{x}(t)=-x$ with $x(0)=\omega \in \mathbb{R}_+$. Since $x(t;\omega)=~\omega e^{-t}$, then $x(t;\omega)\approx 0$ for $t\gg 1$ independently of $\omega$. Since the kernels we use in Algorithms~\ref{alg:search}~and~\ref{alg:opt} weight all times equally, the longer the signal is, the more weight that is given to times $t\gg 1$ where $\omega^*$ is practically non-identifiable; see discussion in Sec.\ \ref{sec:degen}.

\subsection{Lorenz '63 system}\label{sec:lorenz}

\begin{figure}[htb!]
\begin{center} 
\includegraphics[width=0.45\textwidth]{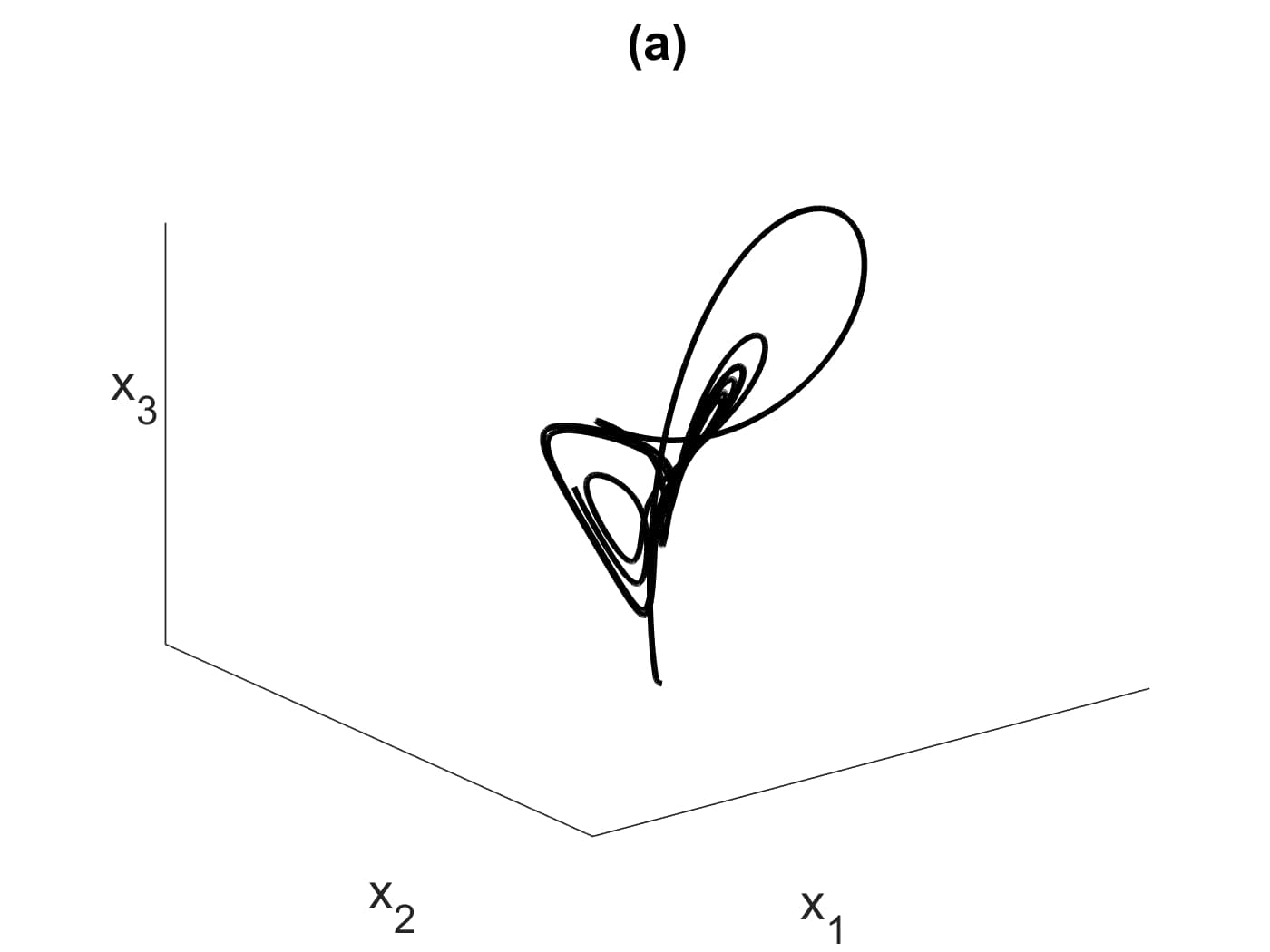}
\includegraphics[width=0.45\textwidth]{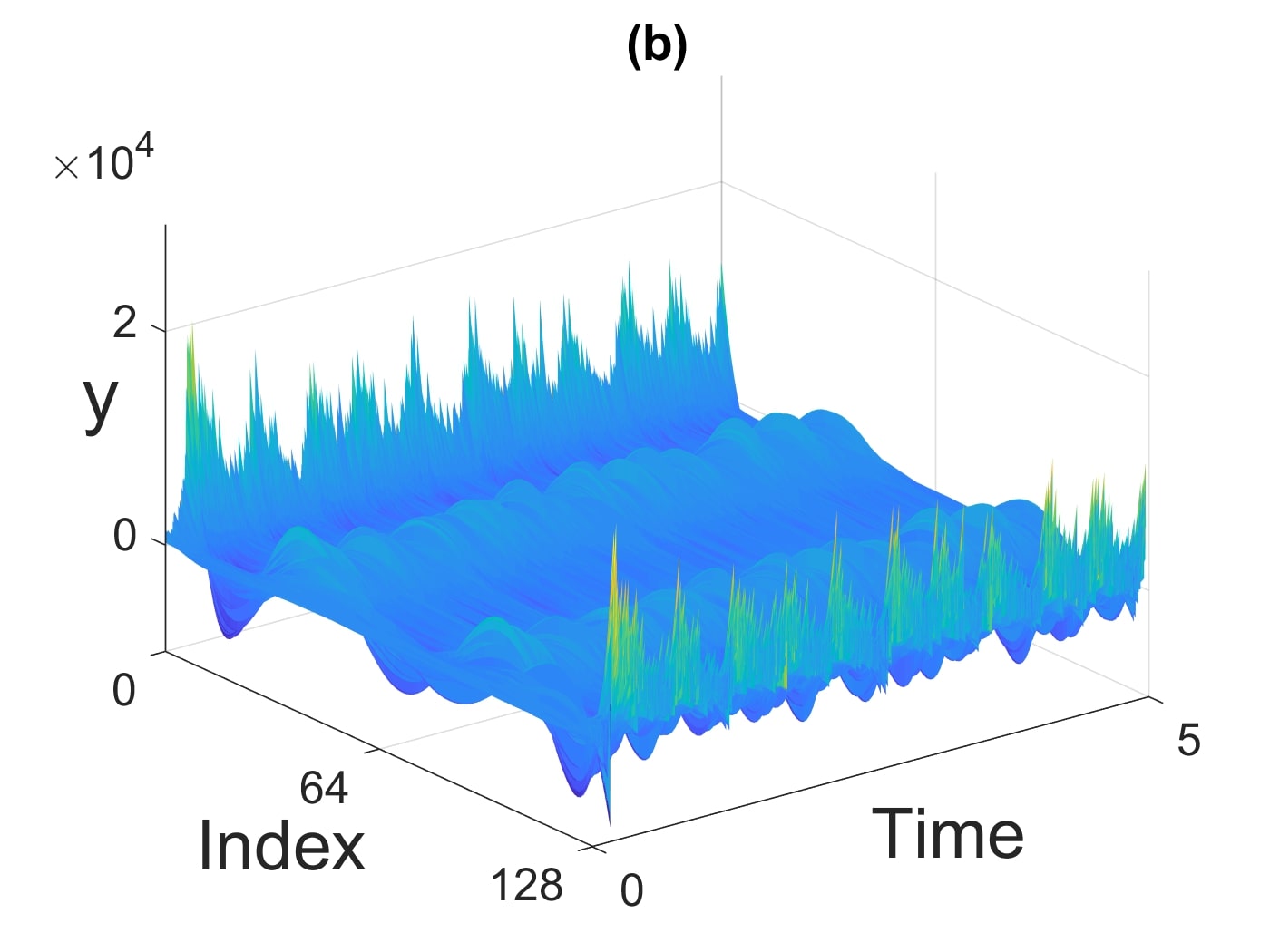}
\includegraphics[width=1\textwidth]{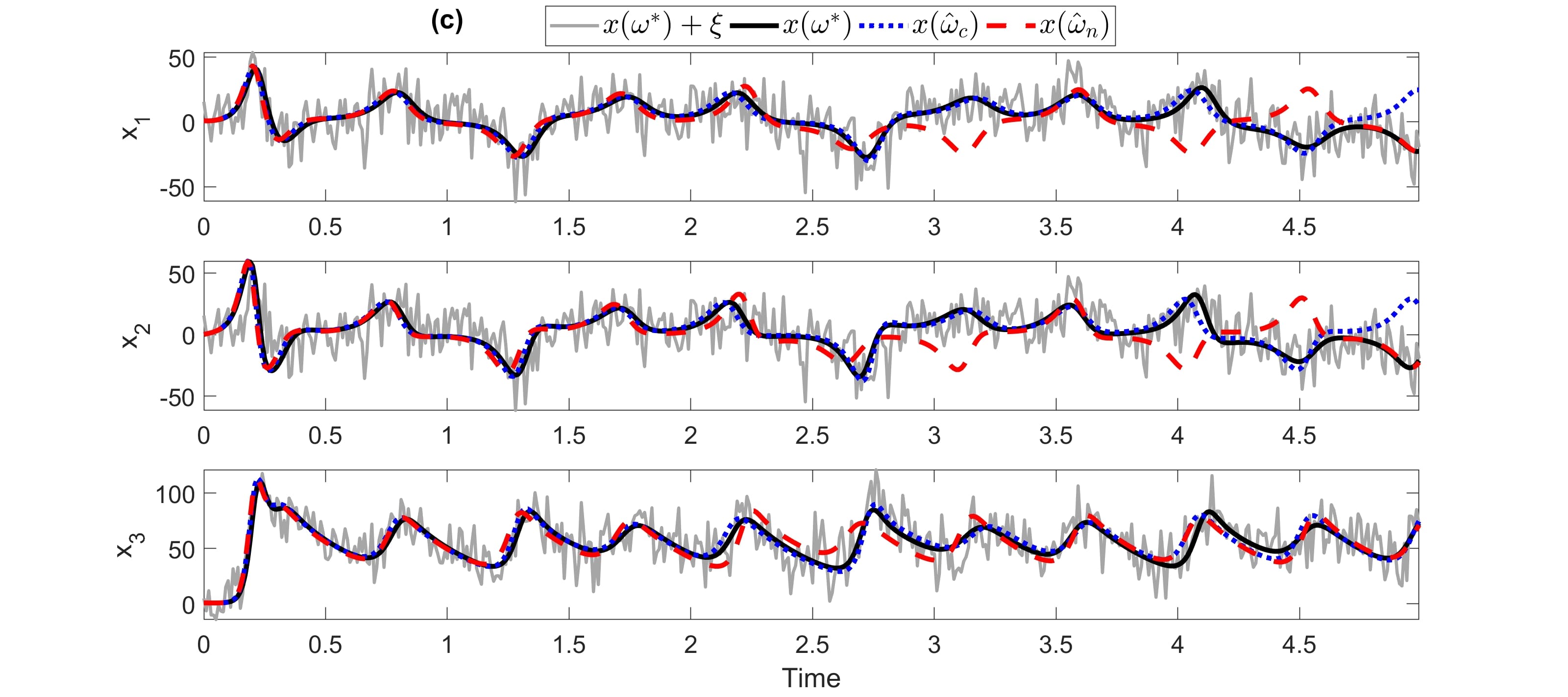}
\end{center}
\caption{(a) Lorenz system \eqref{eq:lorenz} with  {\rev $\rho =60 $, $\sigma = 20 $}, $\beta =~8/3 $. (b) The observed $y(t)$, see \eqref{eq:lorenz_high}. (c) The coordinates of $x(t;\omega^*)$ (solid, black) and its noisy variant $x(t;\omega^*)+\zeta$ (grey) vs. time for the true parameter. For each coordinate we present the trajectory using the estimated parameters based on the clean signal $\hat{\omega}_c$ (dots, blue) and noisy signal $\hat{\omega}_n$. (dashes, red)}
\label{fig:lorenz_example}
\end{figure}

Consider the Lorenz (or Lorenz '63 \cite{lorenz63}) System of ODEs
\begin{equation}
\label{eq:lorenz}
\begin{split}
\dot{x}_1 (t) &=\sigma(x_2-x_1) \, ,\\
\dot{x}_2 (t) &=x_1(\rho -x_3)-x_2 \, ,\\
\dot{x}_3 (t) &=x_1x_2 -\beta x_3 \, ,
\end{split}
\end{equation}
where $\sigma, \rho, \beta \in \mathbb{R}_+^3$ are the model parameters. The Lorenz System is nonlinear, and for certain parameters and initial conditions it is chaotic; see Fig.~\ref{fig:lorenz_example}(a). To embed this system in a high-dimensional space, we follow the nonlinear transformation introduced in \cite{kutz2016pnas, kutz2019data}.  Let $u_j \in \mathbb{R}^{128}$ be the $j$-th order Legendre polynomial evaluated on $128$ uniformly-spaced points in $[-1,1]$, and let\footnote{This precise form of $G$ is not particularly important, only that it is truly nonlinear and incorporates all of $x$'s coordinates. Other observation functions were tested to yield similar result with our algorithm (results not shown).}
\begin{equation}
\label{eq:lorenz_high}
y(t) = G(x(t)) :\,= u_1 {x}_1(t)+ u_2 {x}_2(t)+ u_3 {x}_3(t)+ u_4 {x}_1(t)^2+ u_5 {x}_2(t)^2+ u_6 {x}_3(t)^2 \, ,
\end{equation} where the $\omega$ notations were omitted for brevity; see Fig.\  \ref{fig:lorenz_example}(b). We apply Algorithm~\ref{alg:opt} to $y(t)$ to estimate $\omega^*$. Our algorithm's estimation of these parameters leads to nearly indistinguishable low-dimensional trajectories $x (t)$; see Fig.\ \ref{fig:lorenz_example}(c). 

We then consider a {\em noisy} observation function
\begin{equation}\label{eq:lorenz_high_noise}
    \mathcal{G}(x(t);\zeta) = G(x(t)+\zeta) \, ,
\end{equation}
where $G$ is given by \eqref{eq:lorenz_high} and $\zeta \sim \mathcal{N}(0, \sigma^2 I_3)$ is three-dimensional normally distributed with $\sigma=15$. Note that in this case, the resulting noise in the observation is neither additive nor Gaussian. The estimated parameters produce comparable trajectories; see Fig.\ \ref{fig:lorenz_example}(d).

\begin{figure}[htb!]
\begin{center} 
\includegraphics[width=0.45\textwidth]{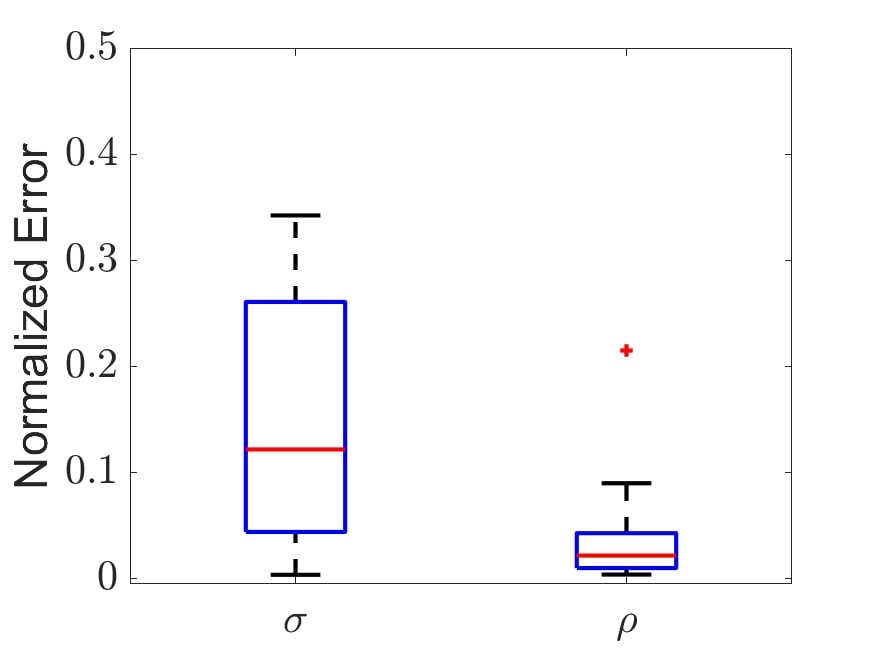}
\includegraphics[width=0.45\textwidth]{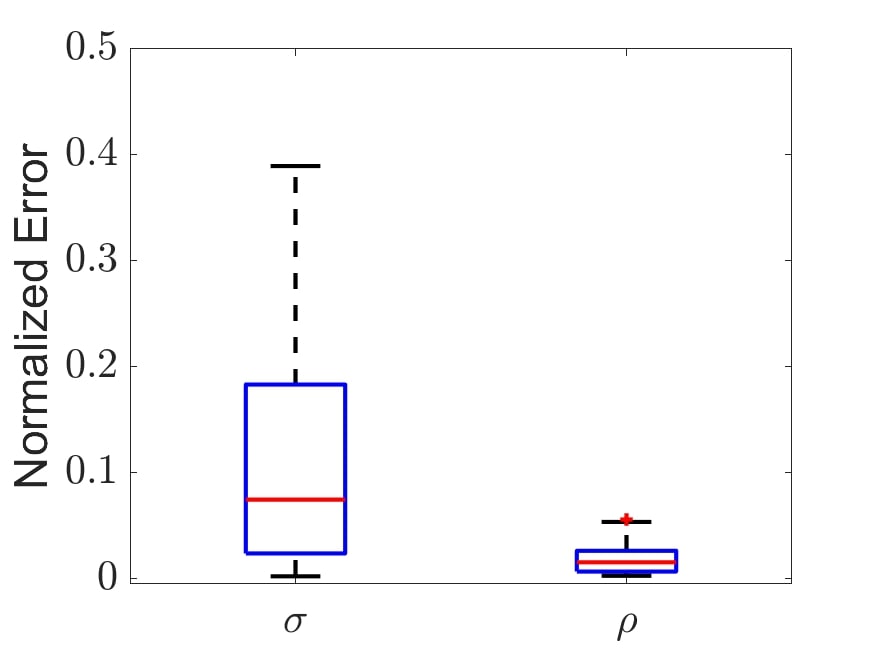}
\includegraphics[width=0.45\textwidth]{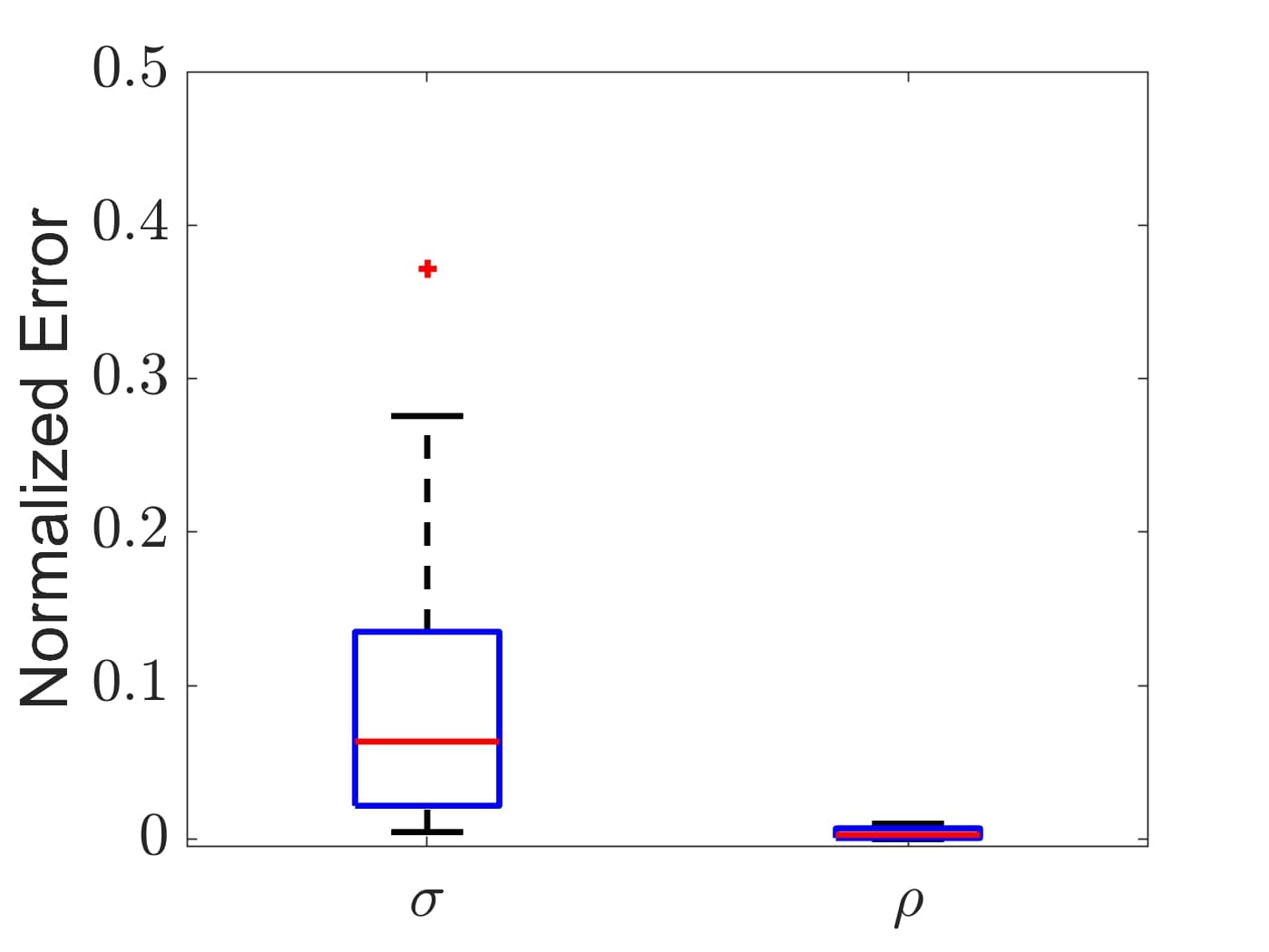}
\includegraphics[width=0.45\textwidth]{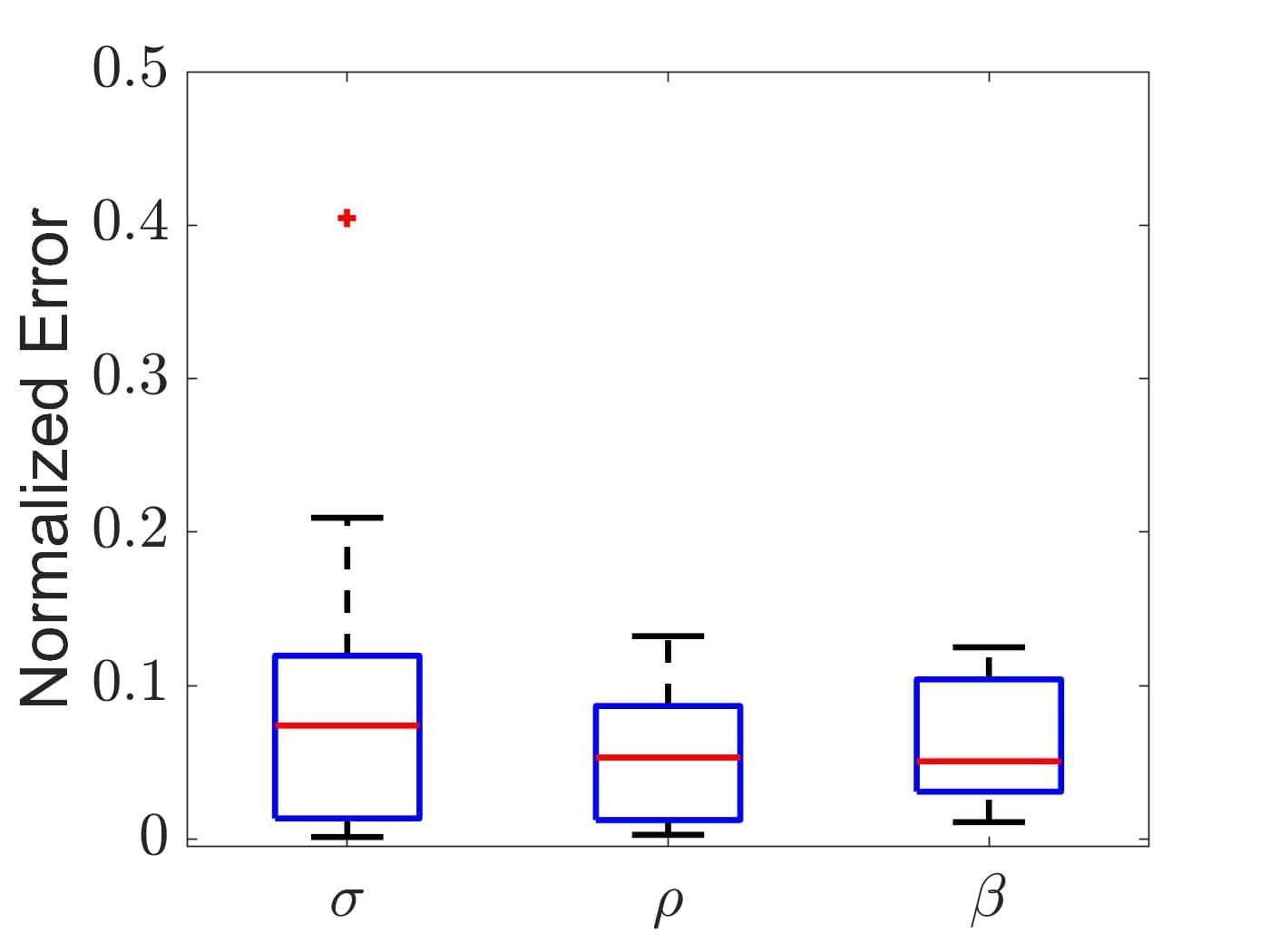}
\end{center}
\caption{Box plots of normalized errors \eqref{eq:estimation_err} for estimated parameters $\sigma$ and $\rho$ for the Lorenz '63 model \eqref{eq:lorenz} given the observation function \eqref{eq:lorenz_high}. Top left: Algorithm \ref{alg:search}. Top right: Algorithm \ref{alg:opt}. Red lines represent the medians and whisker bars indicate the $25$-th and $75$-th quantiles. Bottom left: Oracle estimator using the unknown observation function $G$. {\rev Bottom right: estimation of $3$ parameters using Algorithm \ref{alg:opt}.}}
\label{fig:lorentz_box_plots}
\end{figure}
 
 To systematically study the performance of our proposed approach, we repeat the following experiment $20$ times: we set $\beta=~8/3$, draw $\sigma \in [15,25]$ and $\rho \in [40,80]$, the initial conditions $x_1(0)$, $x_2(0)$, and $x_3[0]$ from $[0,1]$, all uniformly at random. For each set of parameters and initial conditions we solve the Lorenz ODE~\eqref{eq:lorenz} and sample the solution with $\Delta t=0.01$ and $T_{\rm f}=N \cdot \Delta t=10$, where $N=1,000$. For each of these $20$ instances we estimate the parameters $\sigma$, and $\rho$ using Algorithms~\ref{alg:search}~and~\ref{alg:opt}. For Algorithm \ref{alg:search} we use a grid with $50$ values of each parameter, i.e., a search-grid of size $| \Omega_{\rm search} |=50^2=2,500$, with $\Delta \sigma =0.2$ and $\Delta \rho =0.8$. The median normalized estimation error \eqref{eq:estimation_err} of Algorithm \ref{alg:search} is $12 \%$, and $1.2 \%$ for $\sigma$, and $\rho$, respectively. We repeat the same experiment for the optimization-based Algorithm~\ref{alg:opt} with $|\Omega_{\rm init}|=100$, where the overall median normalized errors are $7.5 \%$, and $1.6 \%$ for $\sigma$, and $\rho$, respectively; see box-plots in Fig.\ \ref{fig:lorentz_box_plots}. To provide a baseline, we further evaluate the performance of an ``Oracle'' estimator. We define the Oracle estimator as the linear maximum likelihood estimator \eqref{eq:ls} when $G$ is known. Using this estimator we find the parameter $\omega \in~\Omega_{\rm search}$ that minimizes the square difference between $y$ and $G(x(\omega))$. The box-plots of the oracle estimation appears in the right side of Fig. \ref{fig:lorentz_box_plots}, the median normalized errors are $6.4\%$ and $0.3\%$ for $\sigma$ and $\rho$, respectively. {\rev Next, we use Algorithm \ref{alg:opt} to estimate all three parameters of the Lorenz system $\sigma,\rho$ and $\beta$. To ensure observability, we use $\beta=8/3$ and draw $\sigma \in [15,25]$ and $\rho \in [40,80]$, the initial conditions $x_1(0)$, $x_2(0)$, and $x_3(0)$ from $[0,1]$, all uniformly at random.\footnote{\rev We observe that for $\beta \neq 8/3$, many different sets of parameters produce nearly indistinguishable trajectories, thus obstructing observability, see discussion in Sec.\ \ref{sec:degen}.}  We use $|\Omega_{init}|=100$ ($\beta$ is initialized by drawing from $[2,3]$), and obtain an overall median normalized of $7.4\%$, $5.3\%$ and $5.1\%$ for $\sigma$, $\rho$ and $\beta$, respectively; see bottom right box-plots in Fig. \ref{fig:lorentz_box_plots}.}

 \begin{figure}[htb!]
\begin{center} 
\includegraphics[width=0.5\textwidth]{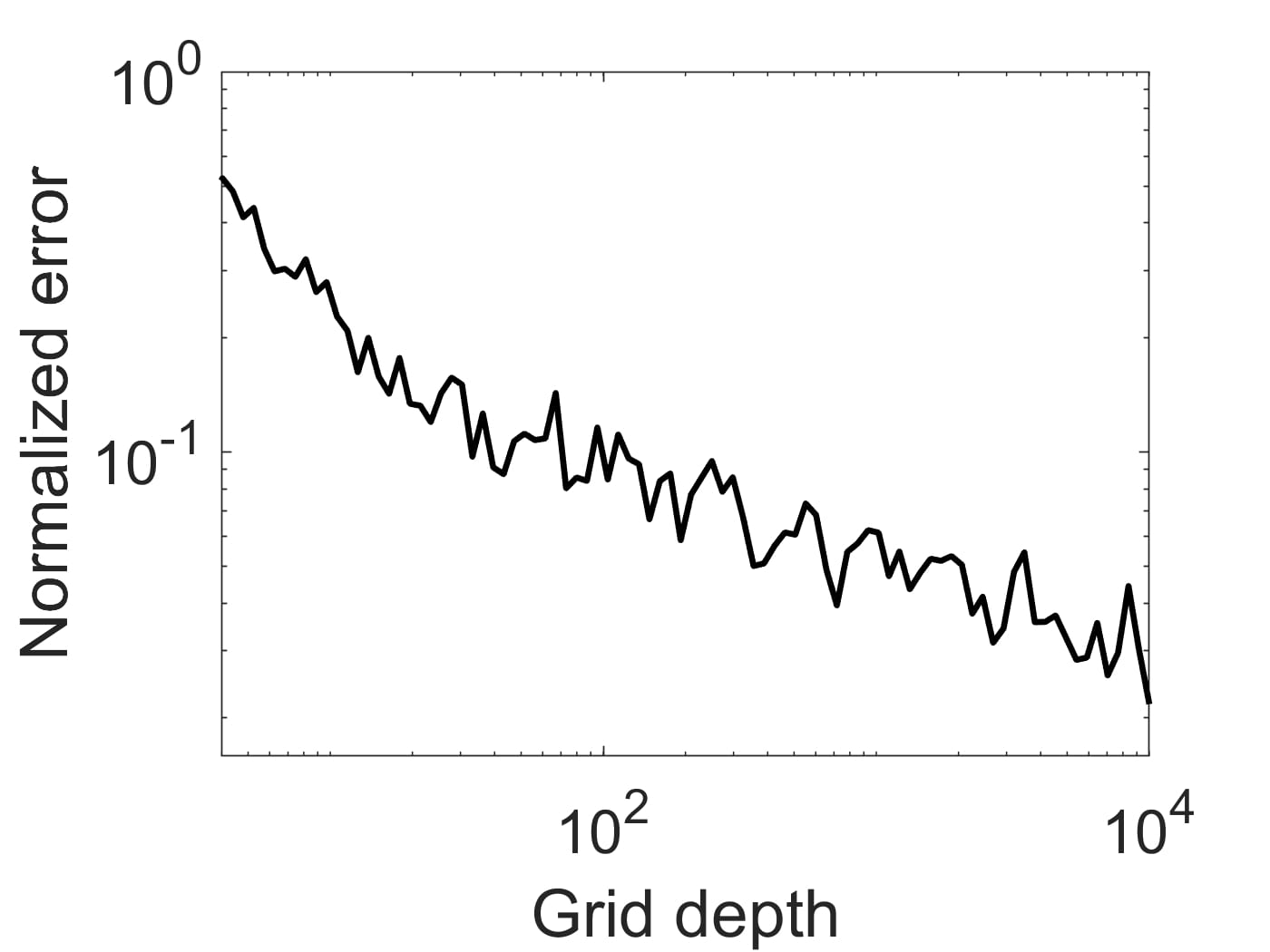}
\end{center}
\caption{Estimating $\sigma$ in the Lorenz equation \eqref{eq:lorenz} from the observations \eqref{eq:lorenz_high} where all other parameters are fixed and known. Parameter estimation error  \eqref{eq:estimation_err} vs. number of grid points $|\Omega_{\rm grid}|$; see Alg. \ref{alg:search}.}
\label{fig:grid}
\end{figure}

  The accuracy of Algorithm \ref{alg:search} is limited by the grid resolution of $\Omega_{\rm search}$. The introduction of an optimization scheme in Algorithm \ref{alg:opt} removes this limitation, but brings a host of other accuracy issues, e.g., the non-convexity of the optimization problem \eqref{eq:w_ml}. But, practical considerations aside, what is the inherent accuracy of the score \eqref{eq:score}, independently of the search or optimization method? To answer this question, we repeated the test of Algorithm \ref{alg:search} on the noiseless observation \eqref{eq:lorenz_high}
of the Lorenz system \eqref{eq:lorenz}, but when {\em only} the parameter $\sigma$ is unknown. This allows us to substantially refine the grid and increase the size of $\Omega_{\rm grid}$ from $10$ to $10^4$. For each grid-resolution we average over $100$ simulations and observe that the median estimation error \eqref{eq:estimation_err} decreases by more than an order of magnitude; see Fig.\ \ref{fig:grid}. Whether this trend saturates at some point, or conversely the error vanishes as $|\Omega_{\rm grid}|\to \infty$, remains an interesting open question.

{\rev 
\subsection{Model coordinates and phase space coordinates}\label{sec:partial}

The results above highlight an important distinction between the model coordinates $x(t;\omega)$ and phase space coordinates. From a dynamical systems perspective, it might seem as if our model coordinates $x(t;\omega)$ in the double pendulum example consist of only partial data - the double pendulum system (Appendix \ref{ap:double}) is a fourth-order ODE, with two angles and two angular velocities, while we use only the position coordinates. Therefore, our model coordinates $x(t;\omega)$ do not specify a single point in phase space. The key observation here is that we observe a time series, and not a single point in time. We claim that temporal derivatives are implied by the time series. This can be understood heuristically, as the differences between subsequent times are indicative of the velocities. In the spirit of Taken's theorem, delay coordinates $(x(t_1;\omega), \ldots , x(t_n;\omega))$ can reconstruct the phase space, especially in this instance where we only ``drop'' two coordinates, see e.g., \cite{dsilva2016data}.

\begin{figure}[htb!]
\begin{center} 

\includegraphics[width=0.6\textwidth]{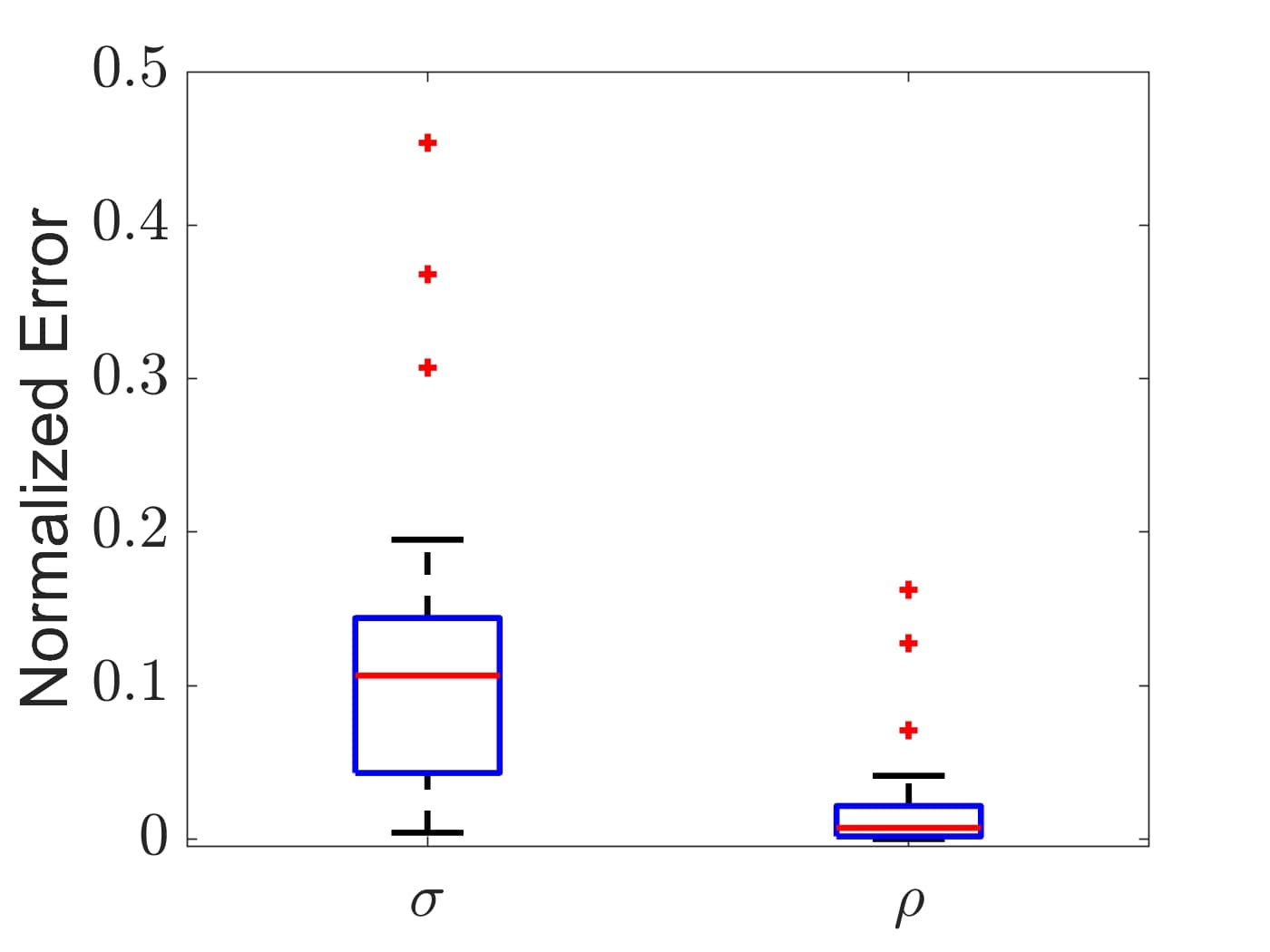}
\end{center}
\caption{\rev Box plots of the normalized errors \eqref{eq:estimation_err} for estimating $\sigma $ and $\rho$ in the Lorenz system \eqref{eq:lorenz} based on the \textit{partial} observations function $y_{\text{partial}}$, see \eqref{eq:partialy}. }
\label{fig:lorentz_partial}
\end{figure}
        
      To test the hypothesis that time-series of partially observed data can be sufficient for parameter estimation purposes, we applied Algorithm \ref{alg:opt} for the case where only two out of the three coordinates of the Lorenz model \eqref{eq:lorenz} are available; we use the following \textit{partial} observation function
      \begin{equation}\label{eq:partialy}
          y_{\text {partial}}(t) = G(x(t)) :\,= u_1 {x}_1(t)+ u_2 {x}_2(t)+ u_4 {x}_1(t)^2+ u_4 {x}_2(t)^2,
          \end{equation}
          where ${x}_1(t)$ and ${x}_2(t)$ are coordinates of the Lorenz '63 system \eqref{eq:lorenz} and $u_i\in \mathbb{R}^{128}$ be the $i$-th Legendre polynomial. Next, we simulate $20$ trajectories based on the same scheme described in section \ref{sec:lorenz}. In this experiment, since $y_{\text {partial}}(t)$ only depends on ${x}_1(t)$ and ${x}_2(t)$, we only used these two coordinates to generate the kernel $K^x$ (see \eqref{eq:kernelx}). We apply Algorithm \ref{alg:opt} with $|\Omega_{init}|=100$, and obtain an overall median normalized of $10.65\%$ and $0.74\%$ for $\sigma$ and $\rho$, respectively; see box-plots in Fig. \ref{fig:lorentz_partial}. This is evidence that a time-series of partially observed data is sufficient to estimate the underlying parameters.
   }

\section{Discussion}\label{sec:discussion}
\subsection{Relevant literature - learning and dynamics}\label{sec:machlearn}

In what follows, we discuss the relation between the main problem of this paper (parameter estimation with an unknown observation function) to notable questions at the interface of dynamical systems, statistics, and machine learning.

 This study is related to the fast-growing field of model discovery and machine learning of physical systems in general \cite{atkinson2019data, bertalan2020learning, bongard2007automated, bramburger2020poincre, dsilva2018chemo, ho2020discover, kutz2016pnas, kutz2019data, yair2017reconstruction}. Generally (notwithstanding their many differences), these works aim to discover governing equations from data using various machine learning techniques. There are three features which distinguish our study from model discovery studies. First, our approach and settings are not agnostic to the physical modelling, and this study does not aim at ``physics discovery" \cite{atkinson2019data, kutz2016pnas}. Rather, we use the laws of physics and known models to estimate the parameters. The second distinction is that the unknown/unspecified portion of our settings is the correspondence between the model and the observations (denoted by $G$, see \eqref{eq:Gdef}), which in model discovery studies is usually assumed to be known. 
  
  The third distinction between this study and machine learning problems in general is that we do not have ample training data, i.e., many pairs of a parameter $\omega$ and the resulting observation $ y(t)$. Nor do we even observe many different signals $y(t)$ which correspond to different (unknown) parameters $\omega$ (as in e.g., \cite{liu2019ml}). Critically, even though one can generate many trajectories $x(t;\omega)$ by solving the underlying ODE, our data includes only a single experiment with a single observed~$y(t)$.
  
  Another increasingly popular application of machine learning to dynamical systems is learning implicit propagation models. By observing many instances of a system's evolution, one learns the time-evolution or Koopman operator to propagate the observations in time {\em without} learning an underlying ODE or PDE, either in a model-free fashion \cite{dubois2020data, pathak2018model, williams2015koopman}, or using partial knowledge of the underlying system \cite{arcomano2020machine, azencot2020forecasting, wikner2020combining, yu2019flowfield}. In this study, propagating the observations $y(t)$ (e.g., a video) in time does not seem to advance the estimation of the system's parameters $\omega^*$. 
  
  Our problem can be considered as a novel variant of standard inverse problems~\cite{aster2018parameter, stuart2010inverse}. Broadly speaking, in these problems a known forward (perhaps noisy) map $\omega \mapsto y(t)=G(x(t;\omega);\zeta)$ is inverted in some sense to recover $x$ or $\omega^*$. This is typically done using the observations $y$ and some a-priori knowledge on the model. The key difference between standard inverse problems and this study is that $G$ in our case is completely unknown, and we do not attempt to recover it. A particularly relevant type of inverse problems is that where $G$ is only known approximately, due to e.g., modelling errors or numerical approximation \cite{cleary2020calibrate, shulkind2018experimental}.
  
Also related to this study is the general problem of nonlinear dimensionality reduction and manifold learning. Since $y=G(x)$ (suppressing noise), then the $\{y(t;\omega)~|~ t\geq 0,~ \omega\in \Omega\}$ can be viewed as a $m+1$ dimensional manifold in the ambient space $\mathbb{R}^D$. Manifold learning techniques can therefore be applied to recover this low-dimensional structure \cite{belkin2003laplacian, coifman2006diffusion, donoho2003hessian, dsilva2018chemo, sober2019manifold,yannis}. These techniques, however, do not provide a straightforward way to compare the sub-manifold resulting from $y(t)$ with the model coordinates $x(t;\omega)$. Even diffusion based techniques which allow one to identify the modality of $x(t;\omega^*)$ in the observation space \cite{lindenbaum2017multi, lederman2018diff} do not lead directly to ways to identify $\omega^*$ from the full parameter space $\Omega$. We consider diffusion-based solutions of our problem an interesting direction of future studies.

\subsection{Future works}
\label{sec:app}

In this work, we presented two examples (the double pendulum and the Lorenz '63 system) where the number of estimated parameters is relatively small. As in other inverse problems, many issues might arise from considering a high-dimensional parameter, e.g., identifiability, computational efficiency, and convergence of the optimization scheme. Adjusting and extending our scheme to high-dimensional parameters is therefore an interesting remaining challenge.

One class of potential applications is the deduction of homogenized constants from reduced models. Consider, for concreteness, a Hamiltonian describing optical propagation of laser beams in a waveguides array, or of a quantum-mechanical electron in two-dimensional graphene. The dynamics of the corresponding system can be approximately reduced to an effective Dirac equation, where short-time dynamics and the micro-structure of the lattice are homogenized to a single constant wave speed. It might be possible, using our method, to observe the full dynamics, and by indirect comparison to the reduced model deduce the homogenized velocity and other constants of the lattice such as topological charges. Even though this problem involves complex PDEs, our proposal is to estimate a low dimensional parameter. This is only an example to a broad class of homogenization schemes in optics, radio-frequency arrays, fluid dynamics, and continuum mechanics. If successful, such an approach could allow measurement from complex dynamical systems by indirect and non-explicit comparison to cheaper and simpler reduced models.  

 Another potential application of our method is predicting hematoma expansion after intracerebral hemorrhage based on non-contrast computed tomography (CT). Over the years pathological observations have led to the development of several parametric models of the propagation of hemorrhage, e.g., \cite{fisher1971pathological, brouwers2012apolipoprotein, brouwers2014predicting}. This is another case where low dimensional parameters underlies high-dimensional data, and requires the ability to estimate the model parameters of a dynamical system from noisy observations consisting of only a single trajectory. Here, the parameter estimation problem is to identify the dynamical regime. One can investigate the ability of our method to estimate the model parameters from a small number of CT scans from a single patient, thereby allowing us to determine whether the hematoma is likely to expand, which in turn necessitates a life-saving, yet risky medical intervention, or whether the hematoma contracts and only monitoring is required. We believe that there exists a large number of such applications from the realm of medical data analysis and related fields, which can benefit from the presented method. 


The approach presented in this paper can be viewed as a general scheme (see Fig. \ref{fig:solution}) of which Algorithms \ref{alg:search} and \ref{alg:opt} are two effective representatives. Our choice of Gaussian kernels in \eqref{eq:kernelx} and \eqref{eq:kernely} is judicious, standard, and effective, but might be improved. One avenue for improvement would be considering kernels that take into account the temporal progression of the samples, as in e.g., \cite{dsilva2018chemo, talmon2013empirical}. One of the main constraints in this paper is that we observe only a single time-series $y(t)$. Suppose the problem is extended to measure many such time-series, either by altering $\omega^*$ or by changing the initial conditions, then an appropriate kernel might be learned, see e.g., \cite{Le,Cortes,owhadi2019kernel}. 

\section{Data Availability Statement}
The data that support the findings of this study are available from the corresponding author
upon reasonable request.

\appendix
{\rev 
\section{Derivation of \eqref{eq:phi2ker} }\label{ap:phi2ker}
We write here the details for $\bar{\Phi}$ and $K^y$.
\begin{align*}
    \|\bar{\Phi}\|_F^2 &= {\rm Tr} (\bar{\Phi} ^{\top} \bar{\Phi}) \\
    &= \sum\limits_{j=1}^N \langle \bar{\Phi}_{\cdot, j}, \bar{\Phi}_{\cdot,j} \rangle \\
    &= \sum\limits_{j=1}^N \langle \phi(y(t_j))- \frac{1}{N} \sum\limits_{i=1}^N \phi (y(t_i)), \phi (y(t_j))- \frac{1}{N} \sum\limits_{i'=1}^N \phi (y(t_{i'})) \rangle \\
    &=\sum\limits_{j=1}^N \left[ \langle \phi(y(t_j)), \phi(y(t_j)) \rangle - \frac{2}{N}\sum\limits_{i}  \langle \phi(y(t_j)), \phi(y(t_i)) \rangle + \frac{1}{N^2} \sum\limits_{i,i'} \langle \phi(y(t_i)), \phi(y(t_{i'})) \rangle \right] \\
    &=\sum\limits_{j=1}^N  \langle \phi(y(t_j)), \phi(y(t_j)) \rangle - \frac{2}{N}\sum\limits_{i,j}  \langle \phi(y(t_j)), \phi(y(t_i)) \rangle + \frac{1}{N^2} \sum\limits_{j,i,i'} \langle \phi(y(t_i)), \phi(y(t_{i'})) \rangle \\
    &=\sum\limits_{j=1}^N  \langle \phi(y(t_j)), \phi(y(t_j)) \rangle - \frac{2}{N}\sum\limits_{i,j}  \langle \phi(y(t_j)), \phi(y(t_i)) \rangle + \frac{N}{N^2} \sum\limits_{i,i'} \langle \phi(y(t_i)), \phi(y(t_{i'})) \rangle \\
    &= \sum\limits_{j=1}^N k^y (y(t_j), y(t_j)) - \frac{1}{N} \sum\limits_{i,\ell} k^y (y(t_i), y(t_{\ell}))\\
    &={\rm Tr}(K^y H) \, , \qquad H_{ij} = \delta_{ij}-\frac{1}{N} \, ,
\end{align*}
where we have used the fact that $\phi$ is real valued and therefore the inner product is symmetric.
}
\section{Proof of Lemma \ref{lem:degeneracy}}\label{ap:normpf}

Since $\alpha \cdot \tr (A) = \tr (\alpha A)$ for every scalar $\alpha$ and square matrix $A$, then  $$s(\utheta) = {\rm Tr} \left( \frac{K^x(\utheta)  H}{\|K^x(\utheta)H\|_F} \frac{K^yH}{\|K^yH\|_F}\right) = \langle \frac{K^x(\utheta)  H}{\|K^x(\utheta)H\|_F} \, , \, \frac{K^yH}{\|K^yH\|_F}\rangle_F  \, , $$ where, as before, $\langle \cdot , \cdot \rangle_F$ is the Frobenius inner product. Hence we can restrict the analysis to the case of $\|K_xH\|_F = \|K_yH \|_F = 1$. We first prove that $$A = \arg \max\limits_{\substack{B\in M_n(\mathbb{R}) \\ \|B\|_F=1}} \langle A, B\rangle_F \, ,$$
for any $A \in M_n(\mathbb{R})$ with $\|A\|_F=1$. Define $\Delta : \, =B-A$. Then 
$$
    1 = \|B\|_F^2 
    = \|A\|_F^2 + \langle A, \Delta \rangle_F + \langle \Delta , A \rangle_F + \|\Delta\|_F^2 \, .$$
Since $A$ and $B$ are real, $\langle A, \Delta \rangle _F = \langle \Delta , A \rangle_F$ and so $\langle A, \Delta \rangle_F = -\|\Delta \|_F^2 /2$. Therefore 
\begin{align*}
     \langle A, B \rangle _F  &=  \langle A, A\rangle_F +\langle A, \Delta \rangle_F \\
     &= \|A\|_F^2 -\frac12 \|\Delta \|_F^2 \\
     &= 1 -\frac12 \|\Delta \|_F^2 \, .
\end{align*}
     Therefore, the maximum of this expression is attained when $\|\Delta \|_F =0$, i.e., when $\Delta = 0$ and $A=B$. 
     
     In the case where $A=K^yH$ and $B=K^x (\omega)H$, the two centered kernels are equal if and only if we have the pairwise equalities $\|y_i(\omega^*)-y_j(\omega^*)\|_2 = \|x_i(\omega)-x_j(\omega)\|_2$ for all times $t_i$ and $t_j$. Since $G$ is an $\ell ^2(\mathbb{R}^d\to \mathbb{R}^d)$ isometry, this inequality occurs if and only if $\|x_i(\omega^*)-x_j(\omega^*)\|_2 = \|x_i(\omega)-x_j(\omega)\|_2$ for all times $t_i$ and $t_j$, i.e., where $x_i(\omega)=Tx_i(\omega ^*)$ for an $\ell^2(\mathbb{R}^d)$ isometry.
     
     \section{Explicit ODEs for the double pendulum}\label{ap:double}
     For completeness, we include here the Euler-Lagrange ODEs that govern the double pednulum system; see \cite{ shinbrot1992chaos} for derivation and details. Denote by $\theta_1$ and $\theta_2$ the angles of the respective pendulums from the negative $y$ axis, then
     $$
    \frac{d}{dt}(\Vec{\theta}) = \frac{d}{dt} \left(\begin{array}{l}
        \theta_1  \\
       \theta _2  \\
         \theta_3 \\
         \theta_4 
    \end{array} \right) = \left(\begin{array}{l}
         \theta_3 \\ \theta_4 \\g_1(\Vec{\theta}) \\ g_1(\Vec{\theta})
    \end{array} \right) \, ,
$$
where $$g_1(\Vec{\theta}) = \frac{g(\sin\theta_2 \cos \Delta \theta -\mu\sin\theta_1)-(l_2\dot{\theta_2}^2+l_1\dot{\theta_1}^2 \cos \Delta \theta )\sin\Delta \theta}{l_1(\mu-\cos^2 \Delta \theta)} \, , $$
$$g_2(\Vec{\theta}) = \frac{g\mu(\sin\theta_1 \cos \Delta \theta -\sin\theta_2)+(\mu l_1 \dot{\theta_1}^2+l_2\dot{\theta_2}^2 \cos \Delta \theta )\sin\Delta \theta}{l_2(\mu-\cos^2 \Delta \theta)} \, , $$
and where
$$\Delta\theta = \theta_1 - \theta _2  \, , \qquad \mu=1+\frac{m_1}{m_2} \, .$$



\begin{thebibliography}{1}
\bibitem{abramovich2013stat}
Abramovich F and Ritov Y,
\newblock{\em Statistical Theory: a Concise Introduction},
\newblock CRC Press, 2013.

\bibitem{arcomano2020machine}
Arcomano T, Szunyogh I, Pathak J, Wikner A, Hunt BR, and Ott E,
\newblock{A machine learning-based global atmospheric forecast model,}
\newblock{Geo. Res. Lett.,} 47, e2020GL087776, 2020.

\bibitem{aster2018parameter}
Aster RC, Borchers B, and Thurber CH,
\newblock {\em Parameter Estimation and Inverse Problems,}
\newblock Elsevier, 2018.

\bibitem{atkinson2019data}
Atkinson S, Subber W, Wang L, Khan G, Hawi P, and Ghanem R,
\newblock Data-driven discovery of free-form governing differential equations,
\newblock {\em arXiv preprint,} arXiv:1910.05117, 2019.

\bibitem{azencot2020forecasting}
Azencot O, Erichson NB, Lin V, and Mahoney MW,
\newblock Forecasting sequential data using consistent Koopman autoencoders,
\newblock{\em arXiv preprint,} arXiv:2003.02236, 2020.

\bibitem{bach2002ica}
Bach FT and Jordan MI,
\newblock Kernel independent component analysis,
\newblock{\em J. Mach. Learn. Res.,} 3, pp.~1--48, 2002.

\bibitem{belkin2003laplacian}
Belkin M and Niyogi P,
\newblock Laplacian eigenmaps for dimensionality reduction and data representation,
\newblock{\em Nueral Comput.,} 15, pp.~1373--1396, 2003.

\bibitem{bertalan2020learning}
Bertalan T, Dietrich F, Mezic I, and Kevrekidis IG.
\newblock On learning Hamiltonian systems from data,
\newblock {\em Chaos,} 29, 121107, 2020.

\bibitem{gretton2008taxonomy}
Blaschko MB and Gretton A. {\em Taxonomy Inference Using Kernel Dependence Measures,} Max-Planck-Insitut Technical Report No. 181, 2008.

\bibitem{bongard2007automated}
Bongard J and Lipson H,
\newblock Automated reverse engineering of nonlinear
dynamical systems,
\newblock {\em Proc. Natl. Acad. Sci.,} 104 pp.~9943--9948, 2007.



\bibitem{boots2012two}
Boots B and Gordon GJ,
\newblock Two-manifold problems with applications to nonlinear system identification,
\newblock {\em Proc. International Conference on Machine Learning,} 2012.



\bibitem{bramburger2020poincre}
Bramburger JJ, and Kutz JN,
\newblock Poincr{\'e} maps for multiscale physics discovery and nonlinear Floquet theory,
\newblock {\em Phys. D,} 408, 132479, 2020.



\bibitem{brouwers2012apolipoprotein}
 Brouwers B, Biffi A, Ayres A M, ,  Schwab K,  Cortellini L, Romero J M, and Goldstein J N.
 \newblock { Apolipoprotein E genotype predicts hematoma expansion in lobar intracerebral hemorrhage.} 
  \newblock {\em Stroke} 43(6), 1490-1495, 2012

\bibitem{brouwers2014predicting}
Brouwers B, Chang Y, Falcone J, Cai X, Ayres M, Battey T, Vashkevich A, McNamara K, Valant V and  Schwab K.
 \newblock Predicting hematoma expansion after primary intracerebral hemorrhage.
  \newblock {\em  JAMA neurology},
  pp. 158--164, 2014.

\bibitem{kutz2016pnas}
Brunton SL, Proctor JL, and Kutz JN. 
\newblock Discovering governing equations from data by sparse identification of nonlinear dynamical systems
\newblock {\em Proc. Natl. Acad. Sci.}, 113:3932--3937, 2016.



 
\bibitem{IPA1}
Byrd RH, Hribar ME, Nocedal J,
\newblock An interior point algorithm for large-scale nonlinear programming,
\newblock{SIAM J. Optim.,} 9, 877-900, 1999.
  
\bibitem{IPA2}
Byrd RH, Gilbert JC, and Nocedal J.
\newblock A trust region method based on interior point techniques for nonlinear programming,
\newblock{\em Math. programming},  89, 149-185, 2000.

\bibitem{kutz2019data}
Champion K, Lusch B, Kutz JN, and Brunton SL. 
\newblock Data-driven discovery of coordinates and governing equations,
\newblock {\em Proc. Natl. Acad. Sci.}, 116:22445--22451, 2019.

\bibitem{high-d}
Charu A, Hinneburg A, and Keim D,
\newblock On the surprising behavior of distance metrics in high dimensional space. \emph{International Conference on Database Theory,} Springer, Berlin, Germany, 2001.

\bibitem{cleary2020calibrate}
Clearey E, Garbuno-Inigo A, Lan S, Schneider T, and Stuart AM,
\newblock Calibrate, emulate, sample,
\newblock{\em arXiv preprint,} arXiv:200.036989.



\bibitem{coifman2006diffusion}
Coifman RR and Lafon S,
\newblock Diffusion maps
\newblock {\em Appl. Comput. Harm. Anal.,} 21, 5--30, 2006.

\bibitem{Cortes}
Cortes C, Mehryar M, and Afshin R,
\newblock Algorithms for learning kernels based on centered alignment,
\newblock{ \em The Journal of Machine Learning Research,} 2019, 13.1, pp.~ 795-828.



\bibitem{ho2020discover}
Cranmer M, Sanchez-Gonzalez A, Battaglia P, Xu R, Cranmer K, Spergel D, and Ho S,
\newblock{Discovering symbolic models from deep learning with inductive biases,}
\newblock{\em arXiv preprint,} arXiv:2006.11287, 2020.

\bibitem{donoho2003hessian}
Donoho DL and Grimes C,
\newblock Hessian eigenmaps: locally linear embedding techniques for high-dimensional data,
\newblock {\em Proc. Natl. Acad. Sci.,} 100, pp.~5591--5596, 2003.

\bibitem{dsilva2018chemo}
Dsilva CJ, Talmon R, Coifman RR, and Kevrekidis IG,
\newblock Parsimonious representation of nonlinear dynamical systems through manifold learning: A chemotaxis case study,
\newblock{\em  Appl. Comput. Harmon. Anal.}, 44, pp.~759--773, 2018.

{\rev
\bibitem{dsilva2016data}
Dsilva CJ, Talmon R, Gear CW, Coifman RR, and Kevrekidis IG,
\newblock Data-driven reduction for a class of multiscale
fast-slow stochastic dynamical systems,
\newblock {\em SIAM J.\ Appl.\ Dyn.\ Sys.}, 15, pp.~1327--1351, 2016.
}
\bibitem{dubois2020data}
Dubois P, Gomez T, Planckaert L, and Perret L.
\newblock Data-driven predictions of the Lorenz system,
\newblock {\em Phys. D,} 132495, 2020.



\bibitem{dunford}
Dunford N and Schwartz JT,
\newblock {\em Linear Operators, Part I},
\newblock Wiley, New York NY, 1958.



\bibitem{fisher1971pathological}
Fisher, C Miller. 
\newblock Pathological observations in hypertensive cerebral hemorrhage. \newblock {\em Journal of Neuropathology \& Experimental Neurology} pp.~536-550, 1971. 



\bibitem{fukumizu2007cca}
Fukumizu K, Bach FR, and Gretton A.
\newblock Statistical Consistency of Kernel Canonical Correlation Analysis.
\newblock {\em J. Mach. Learn. Res.}, 8:361--383, 2007.

\bibitem{fukumizu2008kernel}
Fukumizu K, Gretton A, Sun X, and Sch{\"o}lkopf B.
\newblock Kernel measures of conditional dependence.
\newblock In {\em Advances in Neural Information Processing}, 2008, pp.~489--496.

\bibitem{gretton2005measure}
Gretton A, Bousquet O, Smola A, and Schölkopf B.
\newblock Measuring statistical dependence with Hilbert-Schmidt norms.
\newblock In {\em International Conference on Algorithmic Learning Theory,} 63--77. Springer, Berlin, 2005.

\bibitem{gretton2005kernel}
Gretton A, Herbrich R, Smola AJ, Bousquet O and Sch{\"o}lkpf B.
\newblock Kernel Methods for Measuring Independence,
\newblock {\em J. Mach. Learn. Res.} 6, 2005, pp.~2075--2129.


\bibitem{gretton2008kernel}
Gretton A, Fukumizo K, Teo CH, Song K, Sch{\"o}lkopf B, and Smola AJ.
\newblock A kernel statistical test of independence,
\newblock In {\em Advances in Neural Information Processing}, 2008, pp.~585--592.


\bibitem{high-d2}
Hinneburg A, Charu A, and Keim D,
\newblock What is the nearest neighbor in high dimensional spaces?
\newblock {\em 26th Internat. Conference on Very Large Databases.} 2000.




\bibitem{hofmann2008kernel}
Hofmann T, Sch{\"o}lkopf B, and Smola AJ.
\newblock Kernel Methods in Machine Learning,
\newblock {\em Ann. Statist.} 36, 2008, pp~1171--1220.

\bibitem{iserles2009numerical}
Iserles A,
\newblock {\em A First Course in the Numerical Analysis of Differential Equations,}
\newblock Cambridge University, UK, 2009.

\bibitem{keller2020discovery}
Keller R and Du Q,
\newblock Discovery of Dynamics Using Linear Multistep Methods,
\newblock {\em arXiv preprint,} arXiv:1912.12728, 2020.

 
  	
  	\bibitem{Keller}
  	S.~Lafon, Y.~Keller, and R.~Coifman, ``Data fusion and multicue data matching
  	by diffusion maps,'' \emph{IEEE Trans. Pattern Anal. Mach. Intell.}, vol. 28
  	no. 11, p. 1784–1797, 2006.
  	
 \bibitem{landau1980mechanics}
 Landau LD and Lifshitz EM,
 \newblock{\em Course of Theoretical Physics, Vol.\ 1, Mechanics,}
 \newblock Pergamon, Oxford, UK, 1980.
 
 \bibitem{Le}
Le L, Hao J, Xie Y, and Priestley J.
\newblock Deep kernel: Learning kernel function from data using deep neural network,
\newblock{ \em Proceedings of the 3rd IEEE/ACM International Conference on Big Data Computing, Applications and Technologies.} 2016.

  \bibitem{lederman2018diff}
  Lederman RR and Talmon R,
  \newblock Learning the geometry of common latent variables using alternating-diffusion,
  \newblock{\em  Appl. Comput. Harm. Anal.,} 44, pp.~509--536, 2018.
  
 \bibitem{epsilon}
Lindenbaum O, Salhov M, Yeredor A, and  Averbuch A, \newblock Gaussian bandwidth selection for manifold learning and classification,
  \emph{Data Mining and Knowledge Discovery},  1-37, 2020.
  
  
  \bibitem{lindenbaum2017multi}
  Lindenbaum O, Yeredor A, Salhov M, and Averbuch A,
  \newblock{Multi-view diffusion maps,}
  \newblock{\em Inf. Fusion,} 55, pp.~127--149, 2020.

  


\bibitem{liu2019ml}
Liu C-H, Tao Y, Hsu D, Du Q, and Billinge SJL,
\newblock Using a machine learning approach to determine
the space group of a structure from the atomic pair
distribution function
\newblock {\em Acta Cryst.,} A75, PP.~633--643, 2019.

\bibitem{lorenz63}
Lorenz EN,
\newblock Deterministic nonperiodic flow,
\newblock {J. Atmos. Sci.,} 20, pp.~130--141, 1963.


\bibitem{michaeli2016cca}
Michaeli T, Wang W, and Livescu K,
\newblock Nonparametric canonical correlation analysis
\newblock {\em International Conference on Machine Learning,} pp.~1967--1976, 2016.



\bibitem{nguyen1982ident}
Nguyen VV and Wood EF,
\newblock Review and unification of linear identifiability concepts,
\newblock{\em SIAM Rev.,} 24, pp.~34--51, 1982.

\bibitem{nocedal2006numerical}
Nocedal J and Wright SJ,
\newblock{\em Numerical Optimization}
\newblock Springer, New York NY, 2006.

\bibitem{owhadi2019kernel}
Owhadi H and Yoo GR,
\newblock Kernel flows: from learning kernels from data into the abyss,
\newblock{\em J. Comput. Phys.,} 389, pp.~22--47, 2019.

\bibitem{pathak2018model}
Pathak J, Hunt B, Girvan M, Lu Z, and Ott E,
\newblock Model-free prediction of large spatiotemporally chaotic systems from data: a reservoir computing approach,
\newblock{\em Phys. Rev. Lett.,} 120, 024102, 2018.

\bibitem{saitohBook}
Saitoh S,
\newblock{\em Theory of Reproducing Kernels and its Applications,}
\newblock Longman
Scientific and Technical, Harlow, UK, 1988.

  
  \bibitem{salhov}
  Salhov M, Lindenbaum O, Aizenbud Y, Silberschatz, A, Shkolnisky, Y, and Averbuch, A. 
  \newblock{Multi-view kernel consensus for data analysis,}   \newblock{\em Applied and Computational Harmonic Analysis,} 49(1), 208-228.

\bibitem{scholkopf1998kernel}
Sch{\"o}lkopf B, Smola A, and M{\"u}ller K-S,
\newblock Nonlinear component analysis as a kernel eigenvalue problem,
\newblock{\em Neural Comput.}, 10, pp.~1299--1319, 1998.

\bibitem{sahwe2004kernel}
Shawe-Taylor J and Cristianini N,
\newblock{\em Kernel Methods for Pattern Analysis,}
\newblock Cambridge University, UK, 2004.








\bibitem{shinbrot1992chaos}
Shinbrot T, Grebogi C, Wisdom J, and Yorke JA,
\newblock Chaos in a double pendulum,
\newblock{\em Amer. J. Phys.,} 60, pp.~491--499. 1992.
 

\bibitem{shulkind2018experimental}
Shulkind G, Horesh L, and Avron H,
\newblock Experimental design for nonparametric correction of misspecified dynamical models,
\newblock{\em SIAM/ASA J. Uncertain. Quant.,} 6, pp.~880--906, 2018.

  \bibitem{Singer}
  A.~Singer, R.~Erban, I.~Kevrekidis, and R.~R. Coifman,
  \newblock Detecting intrinsic
  slow variables in stochastic dynamical systems by anisotropic diffusion
  maps,
  \newblock {\em Proc. Nat. Acad. Sci.}, 106, pp. 16090--16095, 2009.
 {\rev 
  \bibitem{smola1998learning}
  Smola AJ and Sch{\"o}lkopf B,
  \newblock {\em Learning with Kernels,}
  \newblock Vol.\ 4, GMD-Forschungszentrum Informationstechnik, 1998.
  }
  \bibitem{sober2019manifold}
Sober B and Levin D,
\newblock Manifold approximation by moving least-squares projection (MMLS),
\newblock {\em Constr. Approx.,} 2019.
  
  \bibitem{stuart2010inverse}
  Stuart AM,
  \newblock Inverse problems: a Bayesian perspective,
  \newblock{\em Acta Numerica,} 19, pp.~451--559, 2010.
  
  \bibitem{talmon2013empirical}
  Talmon R and Coifman RR,
  \newblock Empirical intrinsic geometry for nonlinear modeling and time series filtering,
  \newblock{\em Proc. Natl. Acad. Sci.,} 110, pp.~12535--12540, 2013.
  
\bibitem{vestner2017product}
Vestner M, Litman R, Rodola A, Bronstein M, Cremers D,
\newblock Product manifold filter: Non-rigid shape correspondence via kernel density estimation in the product space,
\newblock {\em Proc. Computer Vision and Pattern Recognition,} 2017.

{\rev 
\bibitem{villaverde2019obs}
Villaverde AF,
\newblock Observability and structural identifiability of nonlinear biological systems,
 \newblock {\em Complexity,} 2019.
}

\bibitem{IPA3}
Waltz RA, Morales JL, Nocedal J, and Orban D,
\newblock An interior algorithm for nonlinear optimization that combines line search and trust region steps,
\newblock{\em Math. Programming,} 107:391--408, 2006.
  	  
\bibitem{FastKNN}
Wang D, Shi L, and  Cao J, 
\newblock Fast algorithm for approximate k-nearest neighbor graph construction.
\newblock In {\em In IEEE 13th International Conference on Data Mining Workshops} pp. 349-356, 2013.




\bibitem{wikner2020combining}
Wikner A, Pathak J, Hunt BR, Girvan M, Arcomano T, Szunyogh I, Pomerance A, and Ott E,
\newblock{Combining machine learning with knowledge-based modeling for scalable forecasting and subgrid-scale closure of large, complex, spatiotemporal systems,}
\newblock {\em Chaos,} MACL2020, 053111, 2020.

\bibitem{williams2015koopman}
Williams MO, Kevrekidis IG, and Rowley CW,
\newblock A data-driven approximation of the Koopman operator: extending dynamic mode decomposition,
\newblock{J. Nonline. Sci.,} 25, pp.~1307--1346, 2015.

\bibitem{yair2017reconstruction}
Yair O, Talmon R, Coifman RR, and Kevrekidis IG,
\newblock Reconstruction of normal forms by learning informed observation geometries from data,
\newblock {\em Proc. Natl. Acad. Sci.,} 114, E7865-E7874, 2017.

\bibitem{yu2019flowfield}
Yu J and Hesthaven JS,
\newblock Flowfield reconstruction method using artificial neural networks,
\newblock{\em AIAA J.,} 57, pp.~482--498, 2019.

\bibitem{zelnik}
  Zelnik-Manor L and Perona P,
  \newblock Self-tuning spectral clustering,
  \newblock in \emph{Advances in Neural Information Processing Systems}, 2004, pp.
  1601--1608.
 
  
\bibitem{yair}
Yair O, Dietrich F, Mulayoff R, Talmon R, Kevrekidis IG,
  \newblock Spectral Discovery of Jointly Smooth Features for Multimodal Data,
  \newblock in \emph{arXiv preprint arXiv:2004.04386}, 2020.
  
  
  \bibitem{yannis}
Holiday A, Kooshkbaghi M, Bello-Rivas JM, Gear CW, Zagaris A, Kevrekidis IG,
 \newblock Manifold learning for parameter reduction, 
 \newblock in \emph{Journal of computational physics}, 2019, 392, pp.419-431.
\end{thebibliography}
\end{document}